\pdfoutput=1

\documentclass[11pt,table]{article}

\usepackage[table,xcdraw]{xcolor}

\usepackage[]{ACL2023}

\usepackage{times}
\usepackage{latexsym}

\usepackage[T1]{fontenc}

\usepackage[utf8]{inputenc}

\usepackage{microtype}

\usepackage{inconsolata}

\usepackage{booktabs}
\usepackage{enumitem}
\usepackage{multirow}
\usepackage{graphicx}

\usepackage[export]{adjustbox}

\usepackage{array} 
\newcolumntype{H}{>{\setbox0=\hbox\bgroup}c<{\egroup}@{}}

\usepackage{pifont}  
\newcommand{\cmark}{\textcolor{green!50!black}{\ding{51}}}
\newcommand{\xmark}{\textcolor{red!70!black}{\ding{55}}}

\usepackage{CJKutf8} 
\newcommand{\chinese}[1]{\begin{CJK*}{UTF8}{gbsn}#1\end{CJK*}}

\usepackage[main=english,russian]{babel}
\newcommand{\russian}[1]{\foreignlanguage{russian}{#1}}

\newcommand{\dalle}{\mbox{DALL$\cdot{}$E}}
\newcommand{\dalletwo}{\mbox{DALL$\cdot{}$E-2}}

\newcolumntype{L}[1]{>{\raggedright\let\newline\\\arraybackslash\hspace{0pt}}m{#1}}
\newcolumntype{C}[1]{@{\hspace{4pt}}>{\columncolor{white}[4pt]\centering\arraybackslash}p{#1}@{\hspace{4pt}}}
\newcolumntype{R}[1]{@{\hspace{4pt}}>{\columncolor{white}[4pt]\raggedleft\arraybackslash}p{#1}@{\hspace{4pt}}}

\title{Character-Aware Models Improve Visual Text Rendering}

\author{\textbf{Rosanne Liu\Thanks{~Equal contribution.} , Dan Garrette\footnotemark[1] , Chitwan Saharia, William Chan, Adam Roberts,} \\
\textbf{Sharan Narang, Irina Blok, RJ Mical, Mohammad Norouzi, Noah Constant}\footnotemark[1] \\
Google Research \\
\texttt{\{rosanneliu,\,dhgarrette,\,nconstant\}@google.com}
}

\begin{document}
\maketitle
\begin{abstract}
Current image generation models struggle to reliably produce well-formed visual text. In this paper, we investigate a key contributing factor: popular text-to-image models lack character-level input features, making it much harder to predict a word's visual makeup as a series of glyphs. To quantify this effect, we conduct a series of experiments comparing character-aware vs.~character-blind text encoders. In the text-only domain, we find that character-aware models provide large gains on a novel spelling task (WikiSpell). Applying our learnings to the visual domain, we train a suite of image generation models, and show that character-aware variants outperform their character-blind counterparts across a range of novel text rendering tasks (our DrawText benchmark). Our models set a much higher state-of-the-art on visual spelling, with $30$+ point accuracy gains over competitors on rare words, despite training on far fewer examples.
\end{abstract}

\section{Introduction}

\begin{figure*}[ht]

\hfill
\begin{minipage}{0.15\textwidth}
\centering
Character-Blind:
\end{minipage}\hfill
\begin{minipage}{0.85\textwidth}
\includegraphics[width=0.23\textwidth]{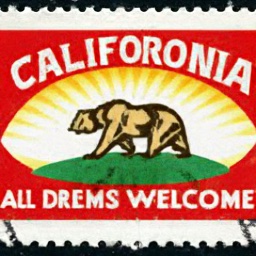}
\includegraphics[width=0.23\textwidth]{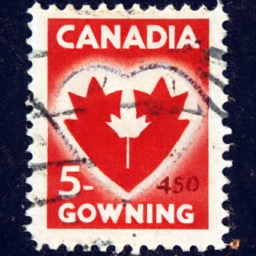}
\includegraphics[width=0.23\textwidth]{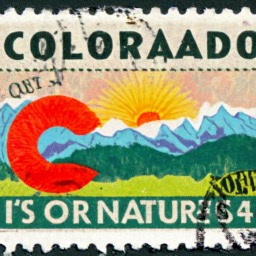}
\includegraphics[width=0.23\textwidth]{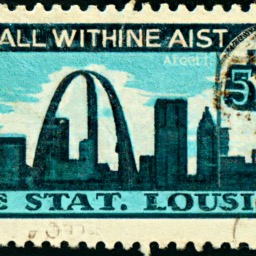}
\end{minipage}

\vspace{1ex}

\hfill
\begin{minipage}{0.15\textwidth}
\centering
Character-Aware:
\end{minipage}\hfill
\begin{minipage}{0.85\textwidth}
\includegraphics[width=0.23\textwidth]{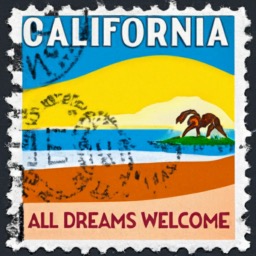}
\includegraphics[width=0.23\textwidth]{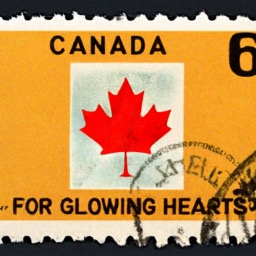}
\includegraphics[width=0.23\textwidth]{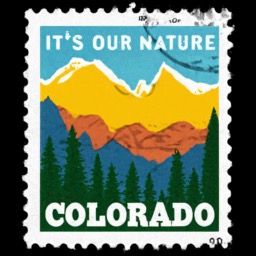}
\includegraphics[width=0.23\textwidth]{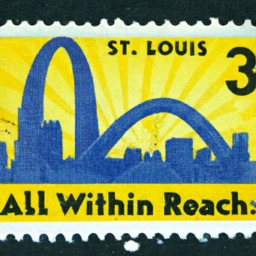}
\end{minipage}

\caption{\textbf{Top}: Image generation models lacking character-level input features often misspell words. \textbf{Bottom}: Using a character-aware text encoder significantly improves the accuracy of rendered text. Prompts are: \textit{A vintage postage stamp with the message:~\rule{1cm}{0.1mm}}, with messages: (1)~\textit{California:~All Dreams Welcome}, (2)~\textit{Canada:~For Glowing Hearts}, (3)~\textit{Colorado:~It's Our Nature}, (4)~\textit{St. Louis:~All Within Reach}.}
\end{figure*}

Over the last year, image generation models have made impressive quality gains \cite{rombach2021high, ramesh2022hierarchical, imagen, parti}. While many practical use cases are already within reach, \emph{rendering visual text} in images remains a challenge. \citet{ramesh2022hierarchical} observe that \dalletwo{} ``struggles at producing coherent text,'' and the latest release of Stable Diffusion lists ``cannot render legible text'' as a known limitation.\footnote{https://huggingface.co/stabilityai/stable-diffusion-2-1}

In this paper, we seek to understand and improve the ability of image generation models to render high-quality visual text. To do so, we first investigate the spelling ability of text encoders in isolation. We find that despite their popularity, \emph{character-blind} text encoders---which receive no direct signal as to the character-level makeup of their inputs---have limited spelling ability. Building on \citet{itzhak-levy-2022-models}, we test the spelling ability of text encoders across scales, architectures, input representations, languages, and tuning methods. We document for the first time the miraculous ability of character-blind models to induce robust spelling knowledge (>$99$\% accuracy) through web pretraining, but show that this does not generalize well beyond English, and is only achieved at scales over 100B parameters, making it infeasible for most applications. We find that \emph{character-aware} text encoders, on the other hand, are able to achieve robust spelling ability at far smaller scales.

Applying these findings to image generation, we train a range of character-aware text-to-image models and demonstrate that they significantly outperform character-blind models on text rendering. For \emph{purely} character-level models, this improved text rendering comes at a cost---decreasing image-text alignment for prompts that don't involve visual text. To alleviate this, we propose combining character-level and token-level input representations, and find that this delivers the best of both worlds.

Our main contributions are to: (1)~Measure the spelling ability of a range of text encoders, pulling apart the effects of scale, character-awareness, and multilinguality, using a new benchmark: WikiSpell. (2)~Present DrawText, the first detailed benchmark of visual text rendering for text-to-image models. (3)~Improve the state of the art in text rendering ability of image generation models through the use of character-aware text encoders.

\section{The spelling miracle}
\label{sec:spelling_miracle}

\begin{figure}
\centering
\includegraphics[width=\columnwidth]{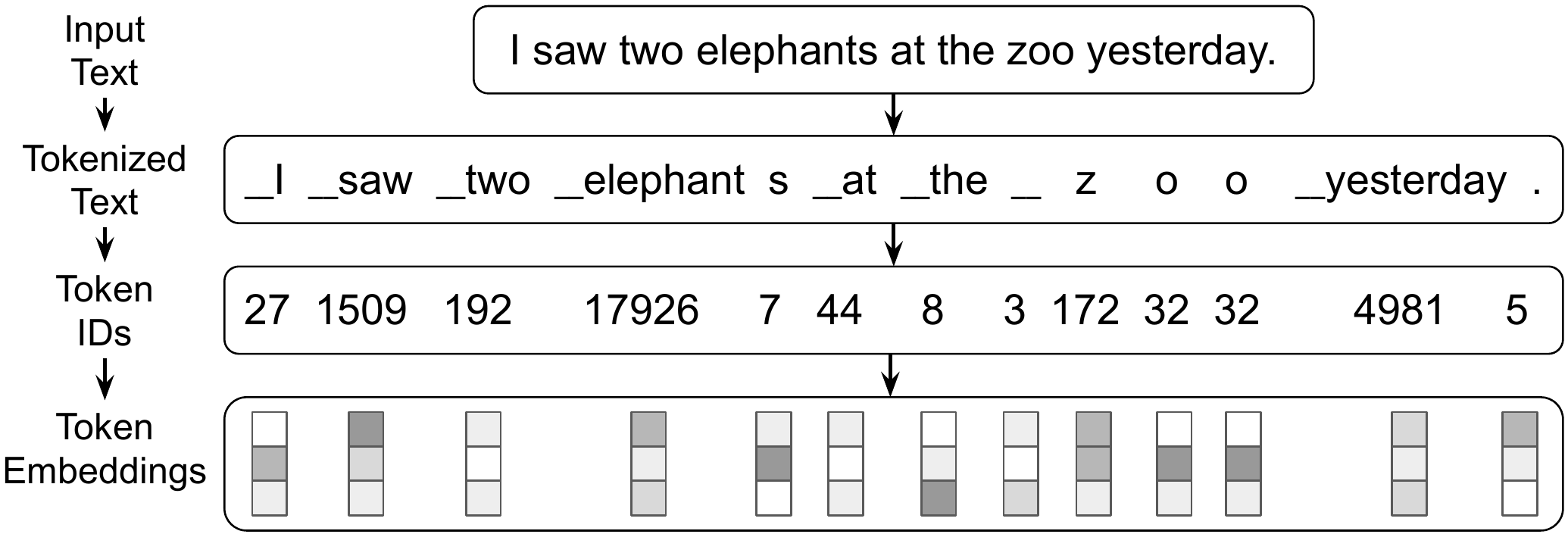}
\caption{Subword tokenization---used in most text-to-image models---maps common character sequences onto IDs that are looked up in an embedding table before being passed to the model, removing any signal about token-internal composition. This example uses the T5 SentencePiece tokenizer \cite{t5}.}
\label{fig:tokenization}
\end{figure}

Language models can be categorized as to whether they have direct access to the characters making up their text input (``character-aware'') or do not (``character-blind''). Many early neural language models operated directly on characters, with no notion of multi-character ``tokens'' \cite{sutskever-2011-generating, graves-2013-generating}. Later models moved to vocabulary-based tokenization, with some like ELMo \cite{peters-etal-2018-deep} retaining character-awareness, and others like BERT \cite{devlin-etal-2019-bert} abandoning it in favor of more efficient pretraining. At present, most widely used language models are character-blind, relying on data-driven subword segmentation algorithms like Byte Pair Encoding (BPE) \cite{gage1994new, sennrich-etal-2016-neural} to induce a vocabulary of subword pieces. While these methods back off to character-level representations for sufficiently \emph{rare} sequences, they compress \emph{common} character sequences into unbreakable units by design, as shown in Figure~\ref{fig:tokenization}.

Recent work on ``token-free'' modeling has pointed to advantages of character-aware input representations. \citet{xue-etal-2022-byt5} show that ByT5---a character-aware multilingual language model trained directly on UTF-8 bytes---outperforms parameter-matched character-blind models on tasks related to spelling and pronunciation. While operating at the byte or character level comes at the cost of training and inference speed, additional work suggests that this can be overcome through downsampling \cite{clark-etal-2022-canine, tay2021charformer}. See \citet{mielke2021between} for a recent overview of tokenization methods and character awareness.

Surprisingly, despite lacking direct access to a token's spelling, character-blind models are, to varying degree, able to \emph{infer} the character-level makeup of their tokens. 
\citet{itzhak-levy-2022-models} observe that, after fine-tuning for spelling, RoBERTa and GPT-2 can achieve $32$\% and $33$\% accuracy at spelling held-out tokens. \citet{kaushal-mahowald-2022-tokens} confirm this ability and probe it further; however it remains unclear where in pretraining this knowledge is coming from, and how to improve it. For example, should we expect larger character-blind models to reach 100\% spelling accuracy across all tokens in their vocabulary?

In section \S\ref{sec:text} we find that, with sufficient scale, character-blind models can achieve near-perfect spelling accuracy. We dub this phenomenon the ``spelling miracle'', to emphasize the difficulty of inferring a token's spelling from its distribution alone. At the same time, we observe that character-blind text encoders of the sizes used in practice for image generation are lacking core spelling knowledge.

With this in mind, it is unsurprising that today's image generation models struggle to translate input tokens into glyph sequences. These models' text encoders are all character-blind, with Stable Diffusion, \dalle{}, \dalletwo{}, Imagen, Parti and eDiff-I all adopting BPE tokenizers \cite{rombach2021high, dalle, ramesh2022hierarchical, imagen, parti, balaji-2022-ediff-i}.

For image-text models, another key source of knowledge is supervised image-caption data. Even if its text encoder is character-blind, could a model learn to spell by observing the makeup of words within images? While possible, we suspect this is an inefficient paradigm for learning, as each token would need to be learned separately, and would need to appear within an image-caption pair seen in training. In section \S\ref{sec:image_experiments} we find that, indeed, this ``late-stage'' learning of spelling is inferior to using a pretrained character-aware text encoder.

\section{Measuring text encoder spelling ability}
\label{sec:text}

Since text-to-image generation models rely on text encoders to produce the representations for decoding, we first explore the ability of text encoders in isolation, using a text-only spelling evaluation task.

\subsection{The WikiSpell benchmark}
\label{sec:wikispell}

We build the WikiSpell benchmark by sampling words from Wiktionary.\footnote{https://www.wiktionary.org/} For each example in the dataset, the input to the model is a single word, and the expected output is its spelling, generated by inserting spaces between each Unicode character:\footnote{i.e., in Python 3: \texttt{" ".join(word)}}
\begin{verbatim}
       elephant  →  e l e p h a n t
\end{verbatim}

To examine the relationship between a word's frequency and a model's ability to spell it, we group words into buckets based on their frequency in the mC4 corpus \cite{xue-etal-2021-mt5}. We create test and development sets from each bucket by sampling $1{,}000$ words uniformly.
The five buckets used are: the top $1$\% most common words, the $1$--$10$\% most common, $10$--$20$\%, $20$--$30$\%, and the bottom $50$\% (which includes words never seen in the corpus).
Finally, we build a training set of $10{,}000$ words by combining $5{,}000$ words sampled uniformly from the bottom $50$\% bucket with $5{,}000$ sampled proportional to their frequencies in mC4.
We exclude words in the dev and test sets from the training set, so evaluation is always on \emph{held out} words.

Beyond English, we repeat this process for six other languages: Arabic, Chinese, Finnish, Korean, Russian, and Thai. For language selection criteria and further technical details, see Appendix~\ref{sec:appendix_wikispell}.

The WikiSpell benchmark is similar to SpellingBee, introduced by \citet{itzhak-levy-2022-models}, but differs in a few key ways. First, SpellingBee is designed to probe a model's \textit{embedding matrix}: given an embedding vector, SpellingBee seeks to output the character sequence of the corresponding vocabulary element. As such, SpellingBee's inputs are always a single token, and it does not measure spelling ability for words the model represents as multiple tokens. Second, due to how subword vocabularies are trained, model vocabularies only contain high-frequency words, and thus SpellingBee inputs are necessarily high-frequency. Finally, as inputs must be drawn from a model's vocabulary, SpellingBee training and evaluation data must be tailored to a specific model, so a dataset cannot be reused across models. In contrast, WikiSpell is model-agnostic, covers single- to many-token words, and covers high- to low-frequency words.

\subsection{Text generation experiments}
\label{sec:text_experiments}

\begin{table*}[h]
\centering
\small
\begin{tabular}{@{}
                L{16mm}
                @{\hspace{1mm}}
                C{3.9mm}
                C{3.9mm}
                C{3.9mm}
                C{3.9mm}
                C{3.9mm}
                C{3.9mm}
                C{3.9mm}
                C{3.9mm}
                @{\hspace{1.5mm}}
                C{3.9mm}
                C{3.9mm}
                C{3.9mm}
                R{3.9mm}
                R{3.9mm}
                R{3.9mm}
                R{3.9mm}
                R{3.9mm}
                @{\hspace{1.5mm}}
                R{3.9mm}
                R{3.9mm}
                R{5.5mm}}
\toprule
                 & \multicolumn{8}{c}{Fine-tuned, with frozen encoder}
                 & \multicolumn{8}{c}{Fine-tuned, all parameters trained}
                 & \multicolumn{3}{c}{Few-shot} \\
                 & \multicolumn{4}{c}{\textbf{T5}}                                              
                 & \multicolumn{4}{c}{\textbf{ByT5}}                                            
                 & \multicolumn{4}{c}{\textbf{T5}}                                              
                 & \multicolumn{4}{c}{\textbf{ByT5}}                                            
                 & \multicolumn{3}{c}{\textbf{PaLM}} \\
\cmidrule(lr){2-5} \cmidrule(lr){6-9} \cmidrule(lr){10-13} \cmidrule(lr){14-17} \cmidrule(lr){18-20}
\multirow{-2}{*}{\shortstack[c]{Frequency\\Bucket}}
    & \textsc{b} & \textsc{l} & \textsc{xl} & \textsc{xxl}
    & \textsc{b} & \textsc{l} & \textsc{xl} & \textsc{xxl}
    & \textsc{b} & \textsc{l} & \textsc{xl} & \textsc{xxl}
    & \textsc{b} & \textsc{l} & \textsc{xl} & \textsc{xxl}
    & \textsc{8B} & \textsc{62B} & \textsc{540B}   \\
\midrule
Top 1\%                            & \cellcolor[HTML]{E67F76}14 & \cellcolor[HTML]{E67C73}12 & \cellcolor[HTML]{F1B7B2}50 & \cellcolor[HTML]{F6D0CD}66 & \cellcolor[HTML]{C8E9D9}97 & \cellcolor[HTML]{FEFEFE}95 & \cellcolor[HTML]{C8E9D9}97 & \cellcolor[HTML]{A2DABE}98 & \cellcolor[HTML]{EDA19B}36 & \cellcolor[HTML]{F0B1AC}46 & \cellcolor[HTML]{F4CAC6}62 & \cellcolor[HTML]{F6D3D0}68 & \cellcolor[HTML]{7DCBA5}99  & \cellcolor[HTML]{57BB8A}100 & \cellcolor[HTML]{57BB8A}100 & \cellcolor[HTML]{57BB8A}100 & \cellcolor[HTML]{FBEDEC}84 & \cellcolor[HTML]{7DCBA5}99 & \cellcolor[HTML]{57BB8A}100 \\
1--10\%                            & \cellcolor[HTML]{EB968F}29 & \cellcolor[HTML]{E98E87}24 & \cellcolor[HTML]{F6D2CF}67 & \cellcolor[HTML]{F7D5D2}69 & \cellcolor[HTML]{C8E9D9}97 & \cellcolor[HTML]{FEFEFE}95 & \cellcolor[HTML]{A2DABE}98 & \cellcolor[HTML]{A2DABE}98 & \cellcolor[HTML]{F6D2CF}67 & \cellcolor[HTML]{F7DAD7}72 & \cellcolor[HTML]{FAE9E8}82 & \cellcolor[HTML]{FBEEED}85 & \cellcolor[HTML]{57BB8A}100 & \cellcolor[HTML]{57BB8A}100 & \cellcolor[HTML]{57BB8A}100 & \cellcolor[HTML]{57BB8A}100 & \cellcolor[HTML]{F4CAC6}62 & \cellcolor[HTML]{A2DABE}98 & \cellcolor[HTML]{7DCBA5}99  \\
10--20\%                           & \cellcolor[HTML]{ECA099}35 & \cellcolor[HTML]{EA938C}27 & \cellcolor[HTML]{F8DBD9}73 & \cellcolor[HTML]{F8DBD9}73 & \cellcolor[HTML]{EDF8F2}96 & \cellcolor[HTML]{FEFCFC}94 & \cellcolor[HTML]{A2DABE}98 & \cellcolor[HTML]{A2DABE}98 & \cellcolor[HTML]{F8DDDA}74 & \cellcolor[HTML]{FAE5E3}79 & \cellcolor[HTML]{FDF4F4}89 & \cellcolor[HTML]{FDF7F7}91 & \cellcolor[HTML]{57BB8A}100 & \cellcolor[HTML]{57BB8A}100 & \cellcolor[HTML]{57BB8A}100 & \cellcolor[HTML]{57BB8A}100 & \cellcolor[HTML]{F7D6D3}70 & \cellcolor[HTML]{C8E9D9}97 & \cellcolor[HTML]{7DCBA5}99  \\
20--30\%                           & \cellcolor[HTML]{EB9B94}32 & \cellcolor[HTML]{E98E87}24 & \cellcolor[HTML]{F6D3D0}68 & \cellcolor[HTML]{F6D3D0}68 & \cellcolor[HTML]{EDF8F2}96 & \cellcolor[HTML]{FEFCFC}94 & \cellcolor[HTML]{7DCBA5}99 & \cellcolor[HTML]{A2DABE}98 & \cellcolor[HTML]{F8DDDA}74 & \cellcolor[HTML]{F9E3E1}78 & \cellcolor[HTML]{FCF1F0}87 & \cellcolor[HTML]{FDF6F5}90 & \cellcolor[HTML]{57BB8A}100 & \cellcolor[HTML]{57BB8A}100 & \cellcolor[HTML]{57BB8A}100 & \cellcolor[HTML]{57BB8A}100 & \cellcolor[HTML]{F7D8D6}71 & \cellcolor[HTML]{C8E9D9}97 & \cellcolor[HTML]{7DCBA5}99  \\
Bottom 50\%                        & \cellcolor[HTML]{EB968F}29 & \cellcolor[HTML]{E88B83}22 & \cellcolor[HTML]{F5CDCA}64 & \cellcolor[HTML]{F5CFCB}65 & \cellcolor[HTML]{C8E9D9}97 & \cellcolor[HTML]{FEFEFE}95 & \cellcolor[HTML]{7DCBA5}99 & \cellcolor[HTML]{A2DABE}98 & \cellcolor[HTML]{F8DEDC}75 & \cellcolor[HTML]{F9E1DF}77 & \cellcolor[HTML]{FCF3F2}88 & \cellcolor[HTML]{FDF6F5}90 & \cellcolor[HTML]{57BB8A}100 & \cellcolor[HTML]{57BB8A}100 & \cellcolor[HTML]{57BB8A}100 & \cellcolor[HTML]{57BB8A}100 & \cellcolor[HTML]{F7D6D3}69 & \cellcolor[HTML]{C8E9D9}97 & \cellcolor[HTML]{7DCBA5}99 \\
\bottomrule
\end{tabular}
\caption{WikiSpell exact-match accuracy results for English. T5 models range from Base (\textsc{b}) (250M params) to XXL (11B params), while ByT5 models range from Base (300M) to XXL (13B).}
\label{tab:wikispell-results-english}
\end{table*}

\begin{table}[]
\centering
\resizebox{\columnwidth}{!}{
\footnotesize
\begin{tabular}{
    @{}
    l
    @{}
    R{4mm}R{4mm}R{4mm}R{4mm}
    @{\hspace{1.5mm}}
    R{4mm}R{4mm}R{4mm}R{4mm}
    @{\hspace{1.5mm}}
    R{4mm}R{4mm}R{5.5mm}
    @{}}
\toprule
                 & \multicolumn{8}{c}{Fine-tuned, with frozen encoder}
                 & \multicolumn{3}{c}{Few-shot} \\
& \multicolumn{4}{c}{\textbf{mT5}}
& \multicolumn{4}{c}{\textbf{ByT5}}
& \multicolumn{3}{c}{\textbf{PaLM}}
\\ 
\cmidrule(lr){2-5} \cmidrule(lr){6-9} \cmidrule(lr){10-12}
Language
    & \textsc{b} & \textsc{l} & \textsc{xl} & \textsc{xxl}
    & \textsc{b} & \textsc{l} & \textsc{xl} & \textsc{xxl}
    & \textsc{8B} & \textsc{62B} & \textsc{540B}   \\
\midrule
Arabic   & \cellcolor[HTML]{EA958E}22 & \cellcolor[HTML]{F7D5D2}60 & \cellcolor[HTML]{FBEEED}75 & \cellcolor[HTML]{EDF8F3}87 & \cellcolor[HTML]{5ABD8C}99 & \cellcolor[HTML]{5DBE8E}99 & \cellcolor[HTML]{59BC8B}100 & \cellcolor[HTML]{5ABD8C}99 & \cellcolor[HTML]{EDA59F}32 & \cellcolor[HTML]{F9E2E0}68 & \cellcolor[HTML]{CEECDD}89 \\
Chinese  & \cellcolor[HTML]{FCF2F1}78 & \cellcolor[HTML]{FCEFEE}76 & \cellcolor[HTML]{FEFBFA}83 & \cellcolor[HTML]{FEFDFD}84 & \cellcolor[HTML]{5DBE8F}99 & \cellcolor[HTML]{66C195}98 & \cellcolor[HTML]{5EBE8F}99  & \cellcolor[HTML]{63C092}99 & \cellcolor[HTML]{FDF8F7}81 & \cellcolor[HTML]{9FD9BD}93 & \cellcolor[HTML]{66C195}98 \\
English  & \cellcolor[HTML]{E67C73}7  & \cellcolor[HTML]{EEA6A0}32 & \cellcolor[HTML]{F5CBC8}54 & \cellcolor[HTML]{FAE6E4}71 & \cellcolor[HTML]{70C59C}98 & \cellcolor[HTML]{7DCBA5}96 & \cellcolor[HTML]{61C091}99  & \cellcolor[HTML]{64C193}99 & \cellcolor[HTML]{FAE7E5}71 & \cellcolor[HTML]{71C69C}97 & \cellcolor[HTML]{60BF91}99 \\
Finnish  & \cellcolor[HTML]{E78178}10 & \cellcolor[HTML]{EFADA7}36 & \cellcolor[HTML]{F7D8D5}62 & \cellcolor[HTML]{FCF1F0}77 & \cellcolor[HTML]{6EC59A}98 & \cellcolor[HTML]{78C9A1}97 & \cellcolor[HTML]{5FBF90}99  & \cellcolor[HTML]{63C092}99 & \cellcolor[HTML]{F2BBB6}45 & \cellcolor[HTML]{FEFDFD}84 & \cellcolor[HTML]{62C092}99 \\
Korean   & \cellcolor[HTML]{EFAFA9}37 & \cellcolor[HTML]{F6D1CE}58 & \cellcolor[HTML]{FCF1F1}77 & \cellcolor[HTML]{FDF8F7}81 & \cellcolor[HTML]{63C092}99 & \cellcolor[HTML]{60BF90}99 & \cellcolor[HTML]{57BB8A}100 & \cellcolor[HTML]{63C092}99 & \cellcolor[HTML]{FAE7E6}71 & \cellcolor[HTML]{D9F0E4}88 & \cellcolor[HTML]{83CDA9}96 \\
Russian  & \cellcolor[HTML]{E67F76}9  & \cellcolor[HTML]{F1B5B0}41 & \cellcolor[HTML]{F6D0CD}57 & \cellcolor[HTML]{FCF0EF}76 & \cellcolor[HTML]{5DBE8E}99 & \cellcolor[HTML]{67C296}98 & \cellcolor[HTML]{5BBD8D}99  & \cellcolor[HTML]{5DBE8F}99 & \cellcolor[HTML]{F0B4AF}41 & \cellcolor[HTML]{F7FCFA}86 & \cellcolor[HTML]{6FC59B}98 \\
Thai     & \cellcolor[HTML]{EDA09A}29 & \cellcolor[HTML]{F1B6B1}42 & \cellcolor[HTML]{F2BDB8}46 & \cellcolor[HTML]{F7D5D2}60 & \cellcolor[HTML]{5ABD8C}99 & \cellcolor[HTML]{61BF91}99 & \cellcolor[HTML]{5ABC8C}99  & \cellcolor[HTML]{5BBD8D}99 & \cellcolor[HTML]{EA958D}22 & \cellcolor[HTML]{F0B1AB}39 & \cellcolor[HTML]{F7D9D7}63 \\
\midrule
\textit{Average}      & \cellcolor[HTML]{EC9E97}27 & \cellcolor[HTML]{F3C3BF}49 & \cellcolor[HTML]{F8DDDB}65 & \cellcolor[HTML]{FCF0EF}77 & \cellcolor[HTML]{62C092}99 & \cellcolor[HTML]{69C397}98 & \cellcolor[HTML]{5CBD8E}99  & \cellcolor[HTML]{60BF90}99 & \cellcolor[HTML]{F4C7C3}52 & \cellcolor[HTML]{FDF5F4}79 & \cellcolor[HTML]{B4E1CB}92 \\
\bottomrule
\end{tabular}}
\caption{WikiSpell exact-match accuracy results on $7$ diverse languages, averaged across all frequency buckets. mT5 and ByT5 models were fine-tuned on the combined training sets of all languages; PaLM was prompted with $20$ in-language examples.}
\label{tab:wikispell-results-all-langs}
\end{table}

We use the WikiSpell benchmark to evaluate pretrained text-only models across a variety of scales. 
In particular, we experiment with: 
\textbf{T5} \cite{t5}, a character-blind encoder-decoder model pretrained on English data; 
\textbf{mT5} \cite{xue-etal-2021-mt5}, which is similar to T5, but pretrained on >$100$ languages; 
\textbf{ByT5} \cite{xue-etal-2022-byt5}, a character-aware version of mT5 that operates directly on UTF-8 byte sequences; and 
\textbf{PaLM} \cite{palm}, a decoder-only model of much larger scale, pretrained predominantly on English.
Experimental results from English-only evaluation are shown in Table \ref{tab:wikispell-results-english}, and multilingual evaluation in Table \ref{tab:wikispell-results-all-langs}.

Our first finding is that character-blind models T5 and mT5 perform much worse on the Top-$1$\% most frequent words. This result may seem counter-intuitive since models typically perform \emph{better} on frequent items; however due to how subword vocabularies are trained, common words are typically represented as atomic tokens (e.g., $87$\% of words in the English Top $1$\% bucket are single-token for T5), thereby obscuring their internal makeup. Scores are a bit higher in the mid-frequency buckets, where words are typically broken into a few common tokens, and lower again in the lowest-frequency bucket, where even the subword tokens may be less frequent. The low spelling accuracies indicate that T5's encoder does not retain sufficient information about the spelling of subwords in its vocabulary.

Secondly, we find that for character-blind models, scale is a key factor in spelling ability.
Both T5 and mT5 improve with scale, but even at XXL size, they are not particularly strong (e.g., T5-XXL's performance on common English words is only $66$\%).
Only when character-blind models reach PaLM's scale do we start to see near-perfect spelling ability: PaLM 540B achieves >$99$\% accuracy across all frequency buckets in English, despite the fact that it sees only $20$ examples in its prompt (as opposed to the $1{,}000$ fine-tuning examples shown to T5).
However, performance is lower on other languages.

Our experiments on ByT5 show that character-aware models have far greater spelling ability. ByT5's performance at Base and Large sizes is only slightly behind XL and XXL (though still in at least the mid-$90$\% range), and the frequency of a word has little effect on ByT5's ability to spell it.
These results far exceed those of (m)T5, and are comparable to the English performance of PaLM, which has >$100\times$ more parameters, and exceed PaLM's performance on other languages. These findings indicate that substantially more character-level information is retained by the ByT5 encoder, and in such a way that it can be retrieved from those frozen parameters as needed for the decoding task.

We also test fine-tuning the full model instead of keeping the encoder frozen (also in Table \ref{tab:wikispell-results-english}). When ByT5's encoder is fine-tuned, performance goes to roughly $100$\% across all scales and frequency buckets.
For T5, the effect of full fine-tuning is more mixed: on rare words, it helps a lot ($65$\%\,$\rightarrow$\,$90$\% for T5-XXL on Bottom $50$\%), while for common words, it has little effect ($66$\%\,$\rightarrow$\,$68$\% on Top $1$\%). This tells us that for words that get broken into smaller pieces (which may be repeated across training examples), the model can memorize spelling information provided during fine-tuning, whereas for single-token words, fine-tuning provides no spelling signal since, by definition, that single token will not appear in the fine-tuning dataset.

\section{The DrawText benchmark}

Evaluating text-to-image models has been an ongoing topic of research, with the development of standard benchmarks from COCO~\cite{lin2014microsoft} to DrawBench~\cite{imagen}, and metrics including FID~\cite{heusel2017gans}, CLIP score~\cite{hessel2021clipscore}, and human preferences~\cite{imagen}. However, there has been a lack of work on text rendering and spelling evaluation. To that end, we present a new benchmark, DrawText, designed to measure the text rendering quality of text-to-image models. 
The benchmark consists of two parts, assessing distinct model capabilities: 1)~DrawText Spelling, which evaluates simple word rendering over a large set of English terms; and 2)~DrawText Creative, which evaluates end-to-end text rendering across diverse visual settings.

\subsection{DrawText Spelling}

To measure the spelling ability of image generation models in a controlled and automatable fashion, we construct 500 prompts by sampling 100 words from each of the English WikiSpell frequency buckets (see \S\ref{sec:wikispell}), and plugging them into the template: \textit{A sign with the word ``\rule{1cm}{0.1mm}'' written on it.} For each prompt, we sample 4 images from the candidate model, and assess them using optical character recognition (OCR)-based metrics.\footnote{For models supporting non-square inference, we use 2:1 aspect ratio, for better visualization. This gave a modest gain in OCR performance in preliminary experiments. We rescale all images to $64\times64$ before running OCR for fair comparison.}

For OCR evaluation, we use the Google Cloud Vision API, which returns all text found within an image, along with bounding box locations.
The DrawText Spelling prompt tends to generate a prominently positioned sign with text, which is relatively simple for off-the-shelf OCR to identify, but if the system returns multiple bounding boxes, we only use the top-most one.
Additionally, since text may be rendered across multiple lines, we post-process the OCR output by removing newline characters that appear within a single bounding box.
Finally, since text on real signs is often written in all capitals, and models often do the same regardless of how the word is written in the prompt, we ignore case when computing the spelling accuracy.

\subsection{DrawText Creative}

\begin{figure*}
\centering
\includegraphics[width=0.24\textwidth]{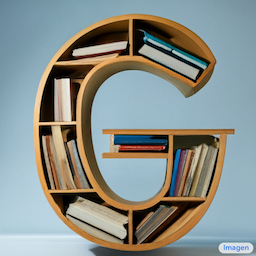}
\includegraphics[width=0.24\textwidth]{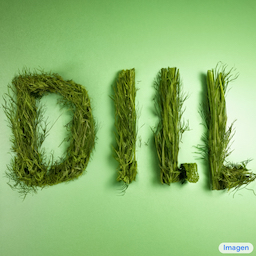}
\includegraphics[width=0.24\textwidth]{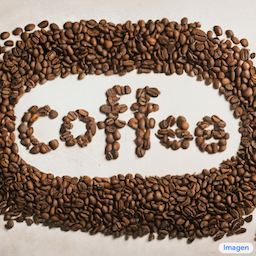}
\includegraphics[width=0.24\textwidth]{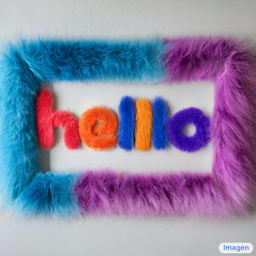}

\vspace{0.15ex}

\includegraphics[width=0.24\textwidth]{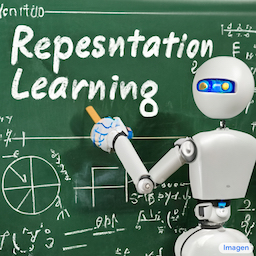}
\includegraphics[width=0.24\textwidth]{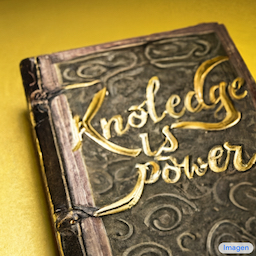}
\includegraphics[width=0.24\textwidth]{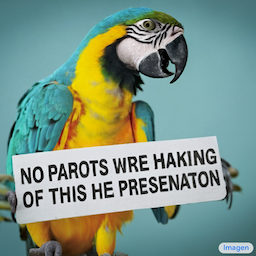}
\includegraphics[width=0.24\textwidth]{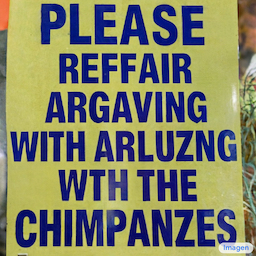}
\caption{Sample generations from Imagen \cite{imagen} on eight DrawText Creative prompts, illustrating their diversity. Most samples exhibit misspelled text or misshapen glyphs (e.g., the \texttt{ff} in \texttt{coffee}). Errors are increasingly common with prompts seeking longer text spans. See Appendix~\ref{sec:appendix_creative_prompts} for all prompts.}
\label{fig:creative_prompts}
\end{figure*}

Visual text is not limited to mundane examples like street signs.  Text can appear in many forms---scribbled, painted, carved, sculpted, and so on.  If image generation models support flexible and accurate text rendering, this can help designers in developing creative fonts, logos, layouts, and more.

To test the ability of image generation models to support these use cases, we worked with a professional graphic designer to construct $175$ diverse prompts that require rendering text in a range of creative styles and settings.  The prompts vary in how much text is specified, ranging from a single letter to an entire sentence. We share these prompts in Appendix~\ref{sec:appendix_creative_prompts}, with the expectation that they will help the community work towards improving text rendering.  Many of the prompts are beyond the abilities of current models, with state-of-the-art models exhibiting misspelled, dropped, or repeated words, as seen in Figure~\ref{fig:creative_prompts}.

\section{Image generation experiments}
\label{sec:image_experiments}

In this section, we evaluate the spelling ability of text-to-image generative models with the proposed DrawText benchmark.  State-of-the-art text-to-image generative models consist of a text encoder plus a cascade of either diffusion models~\cite{imagen} or autoregressive models~\cite{parti} that map the encoded text representations to realistic images.
In section \S\ref{sec:text} we saw that character-aware text encoders greatly outperform character-blind models on spelling in a text-only setting; in this section, we investigate whether making the text encoder character-aware improves the text rendering ability of text-to-image models.

\subsection{Models}

For an apples-to-apples comparison, we train two character-blind and three character-aware image generation models. Our training closely follows the procedure of \citet{imagen}, with the following modifications. First, our models train for $500{,}000$ steps, which is $5.6\times$ fewer steps than Imagen. Second, we only train the initial \mbox{$64\times64$} model, as text rendering ability can already be assessed at this scale. This allows us to forgo the training of super-resolution models.

Third, rather than a mixture of datasets, we train exclusively on the publicly available Laion-400M \cite{laion400m}. This improves reproducibility and also increases the amount of visual text seen during training. Inspecting a random sample of 100 images, we found that a relatively high proportion (around 71\%) of Laion images contain text, and many (around 60\%) exhibit correspondence between caption text and visual text.

Fourth, to prevent models from clipping text, we train on \emph{uncropped} images with arbitrary aspect ratios. In contrast with the widely used strategy of cropping a square from the center of the image, we maintain the image's true aspect ratio by padding with black borders. The model receives an additional binary mask input indicating the padding.\footnote{We apply the above strategy for 80\% of training examples, and use center cropping for the remaining 20\%.}

To test the effects of text encoder size and character-awareness, we vary the pretrained text encoder as follows:

\textbf{T5-XL and T5-XXL} --- Following \citet{imagen}, we use the (character-blind) pretrained T5 text encoders of \citet{t5}. The encoder sizes are 1.2B (XL) and 4.6B (XXL). Note, T5-XXL is the same encoder used in both Imagen and the recent eDiff-I \cite{balaji-2022-ediff-i}.

\textbf{ByT5-XL and ByT5-XXL} --- We use the pretrained ByT5 encoders of \citet{xue-etal-2022-byt5}, with encoders sizes 2.6B (XL) and 9.0B (XXL). These differ from T5 in several regards. First, ByT5 models read and write UTF-8 bytes rather than tokens from a vocabulary, so they are fully character-aware. Second, ByT5 is multilingual, trained on the mC4 corpus of over $100$ languages. Third, ByT5 pretrains with sequence length $1024$, twice that of T5. When encoding text as input to the image generation module, we use a sequence length of $256$ bytes, compared to $64$ tokens for the T5 models.

\textbf{Concat(T5-XXL, ByT5-Small)} --- 
We use as the text encoding a concatenation of the encodings from T5-XXL and a small ByT5 model. ByT5-Small (220M) represents a lightweight addition to the \citet{imagen} model in terms of overall compute and model size (a 4.8\% increase in encoder size), but makes the model character-aware.

\textbf{Imagen Aspect-Ratio (Imagen-AR)} --- To test the benefit of training on uncropped images, we fine-tune the Imagen model of \citet{imagen} for an additional $380{,}000$ steps, to 3.2M steps total, training on uncropped images with preserved original aspect ratio, as described above.

Beyond these custom models, we benchmark Stable Diffusion version 1.5 \cite{rombach2021high}, \dalletwo{} \cite{ramesh2022hierarchical}, Parti \cite{parti} and Imagen \cite{imagen}, all of which use character-blind text encoders. Among these, Imagen is most similar to our experimental models, using the same T5-XXL encoder, but trained much longer and with a larger scale of data.

\subsection{DrawText Spelling results}

\begin{figure*}
\centering
\includegraphics[width=0.85\textwidth, trim={1.5ex, 1.5ex, 1.5ex, 2ex}, clip]{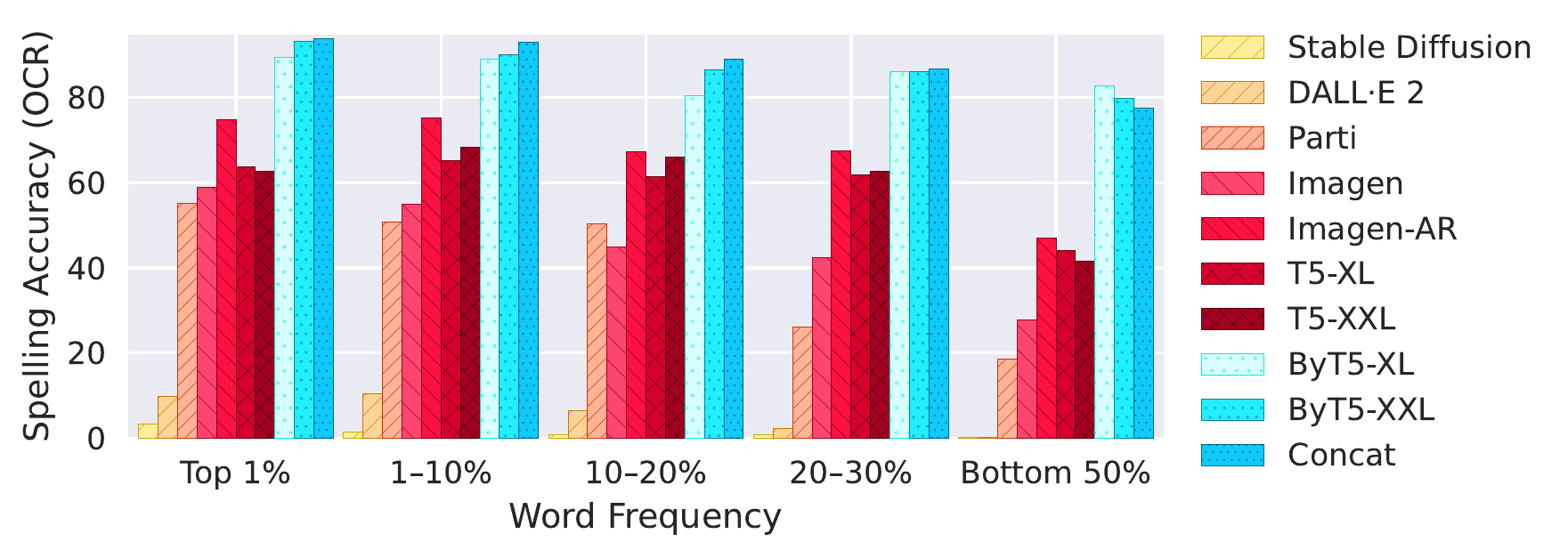}
\caption{Accuracy of $10$ image generation models on DrawText Spelling. Character-aware models (ByT5 and Concat) outperform others regardless of size, and particularly on rare words. Imagen-AR shows the benefit of avoiding cropping, but still underperforms character-aware models, despite training $6.6\times$ longer.}
\label{fig:ocr_accuracy}
\end{figure*}

\paragraph{Char-aware models improve spelling.}
Figure~\ref{fig:ocr_accuracy} shows our DrawText Spelling results across $10$ models, with $2{,}000$ images sampled per model, evaluated with OCR\@. 
Accuracy is computed on the full string (i.e., no credit is given for partial matches).
Across all word frequencies, character-aware models (ByT5 and Concat) outperform the rest, with $15$+ point accuracy gains over Imagen-AR on the most frequent words, and $30$+ point gains on the least frequent words. This is remarkable given that Imagen-AR trained for $6.6\times$ longer.

Our T5 models (character-blind) provide a more controlled comparison against the character-aware models, as they differ \emph{only} in the choice of text encoder---trained on the same dataset for the same number of steps. Here, the gains are even larger: $25$+ point gains on the most frequent words and $30$+ point gains on the least frequent. Notably, these gains persist even for the \emph{smaller} ByT5-XL model, whose encoder is $43$\% smaller than T5-XXL.

To estimate the rate of OCR errors, we manually validate a balanced set of $128$ samples from T5-XXL and ByT5-XXL (see Appendix~\ref{sec:appendix_ocr_validate} for details). We find no false positives, but when OCR detects an error, ByT5-XXL is actually correct $34\%$ of the time, while T5-XXL is correct $9$\%. This asymmetry suggests the benefit of character-aware modeling may be even greater than implied by Figure~\ref{fig:ocr_accuracy}.

\begin{table}[ht]
\centering
\resizebox{0.95\columnwidth}{!}{
\footnotesize
\begin{tabular}{@{}c@{}c@{}HH}
\toprule
\textbf{Error Type} & \textbf{Examples} & \textbf{Char-Blind} & \textbf{Char-Aware} \\

\midrule

Semantic
& \texttt{demonstrated} $\rightarrow$ \texttt{demonstrafied}
& \multirow{2}{*}{\xmark}
& \multirow{2}{*}{\cmark} \\
\cmark & \texttt{inquisitiveness} $\rightarrow$ \texttt{inquisioness} \\

\midrule

Homophone
& \texttt{accommodate} $\rightarrow$ \texttt{accomidate}
& \multirow{2}{*}{\xmark}
& \multirow{2}{*}{\cmark} \\
\cmark & \texttt{Toronto} $\rightarrow$ \texttt{Torondo} \\

\midrule

Add Glyph
& \texttt{labor} $\rightarrow$ \texttt{labort}
& \multirow{2}{*}{\xmark}
& \multirow{2}{*}{\cmark} \\
\cmark & \texttt{debut} $\rightarrow$ \texttt{debust} \\

\midrule

Drop Glyph
& \texttt{stopping} $\rightarrow$ \texttt{stoping}
& \multirow{2}{*}{\xmark}
& \multirow{2}{*}{\xmark} \\
\xmark & \texttt{experiments} $\rightarrow$ \texttt{experimets} \\

\midrule

Repeat Glyph
& \texttt{possible} $\rightarrow$ \texttt{posssible}
& \multirow{2}{*}{\xmark}
& \multirow{2}{*}{\xmark} \\
\xmark & \texttt{locate} $\rightarrow$ \texttt{locaate} \\

\midrule

Merge Glyphs
& \multirow{2}{*}{
  \parbox[c]{0.7in}{\includegraphics[width=0.7in, trim={30pt, 15pt, 30pt, 15pt}, clip]{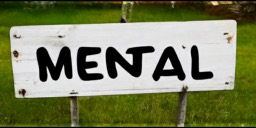}}
  \parbox[c]{0.7in}{\includegraphics[width=0.7in, trim={24pt, 18pt, 24pt, 6pt}, clip]{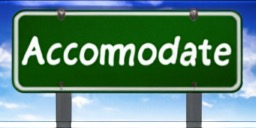}}}
& \xmark & \xmark \\
\xmark \\[1.5ex]

\midrule

Misshape
& \multirow{2}{*}{
  \parbox[c]{0.7in}{\includegraphics[width=0.7in, trim={16pt, 8pt, 16pt, 8pt}, clip]{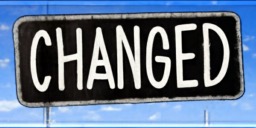}}
  \parbox[c]{0.7in}{\includegraphics[width=0.7in, trim={14pt, 7pt, 14pt, 7pt}, clip]{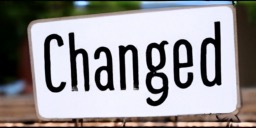}}}
& \xmark & \xmark \\
\xmark \\[1.5ex]

\midrule

No Text
& \multirow{2}{*}{
  \parbox[c]{0.7in}{\includegraphics[width=0.7in, trim={12pt, 0pt, 12pt, 12pt}, clip]{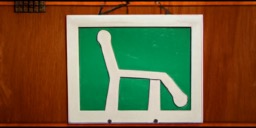}}
  \parbox[c]{0.7in}{\includegraphics[width=0.7in, trim={12pt, 12pt, 12pt, 0pt}, clip]{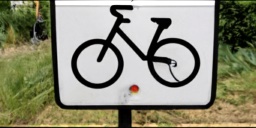}}}
& \xmark & \xmark \\
\xmark \\[1.5ex]

\bottomrule

\end{tabular}}
\caption{Error types observed on DrawText Spelling. Some errors (\cmark) are not observed in character-aware models. Others (\xmark) are found across all model types.}
\label{tab:error_types}
\end{table}

\paragraph{Char-aware models make fewer types of error.}
To gain a better understanding of different models' failure modes, we manually inspect our T5 and ByT5 model outputs. Table~\ref{tab:error_types} illustrates common error types.
Several categories of error are \emph{only} observed in T5 models, suggesting that they stem from the encoder's lack of core spelling knowledge. In \textbf{semantic} errors, the model makes a plausible morpheme substitution, as in \texttt{demonstrated} $\rightarrow$ \texttt{demonstrafied}. In \textbf{homophone} errors, the model produces an incorrect spelling that could be pronounced similarly to the target word. This suggests that some of T5's ``miraculous'' spelling ability may derive from online pronunciation guides. In \textbf{add glyph} errors, the model inserts a letter that was absent from the target, again reflecting the model's uncertainty about a token's internal character makeup.

\begin{figure}
\centering
\includegraphics[width=0.24\columnwidth]{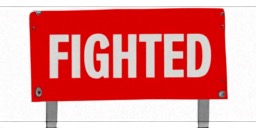}
\includegraphics[width=0.24\columnwidth]{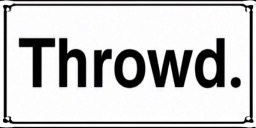}
\includegraphics[width=0.24\columnwidth]{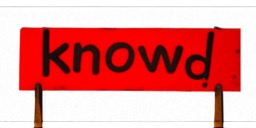}
\includegraphics[width=0.24\columnwidth]{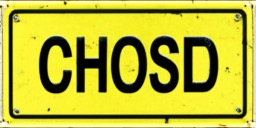}
\caption{Character-blind T5 models erroneously use \texttt{-(e)d} endings for irregular past tense verbs (targets: \textit{fought}, \textit{threw}, \textit{knew}, \textit{chose}).}
\label{fig:irreg-errors}
\vspace{-6px}
\end{figure}

One notable sub-type of semantic error is character-blind models ``regularizing'' irregular inflection, as in \texttt{fought}\,$\rightarrow$\,\texttt{fighted}. On a hand-chosen set of $23$ common irregular past-tense verbs (\textit{began}, \textit{chose}, \textit{dug}, etc.), we find T5-based models erroneously add \texttt{-ed} in 11\% of samples (see Figure~\ref{fig:irreg-errors}), while our character-aware models never exhibit this type of error. This is clear evidence that character-blind models partly rely on a word's meaning (\texttt{fought}\,$\Rightarrow$\,\textsc{past}) and fallible patterns of morphology (\textsc{past}\,$\Rightarrow$\,\texttt{-ed}) to predict spelling.

Other error categories are found across all model types; these include dropped, repeated, merged, or misshapen glyphs. Given that our ByT5 encoders provide a robust spelling signal (see \S\ref{sec:text_experiments}), we understand these errors to be ``layout issues'', where the image generation module has trouble shaping and positioning realistic glyphs within the image.

\begin{figure}
\centering

T5-XXL \\[1ex]

\includegraphics[width=0.24\columnwidth]{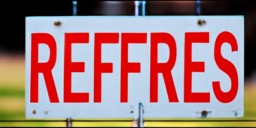}
\includegraphics[width=0.24\columnwidth]{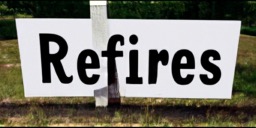}
\includegraphics[width=0.24\columnwidth]{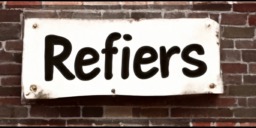}
\includegraphics[width=0.24\columnwidth]{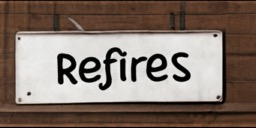}

\includegraphics[width=0.24\columnwidth]{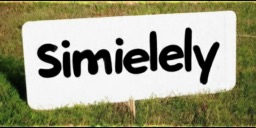}
\includegraphics[width=0.24\columnwidth]{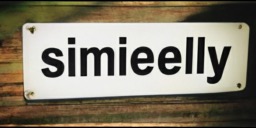}
\includegraphics[width=0.24\columnwidth]{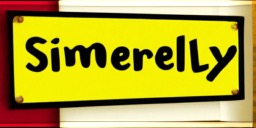}
\includegraphics[width=0.24\columnwidth]{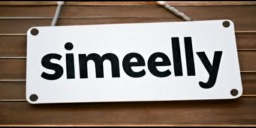}

\includegraphics[width=0.24\columnwidth]{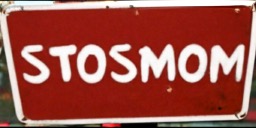}
\includegraphics[width=0.24\columnwidth]{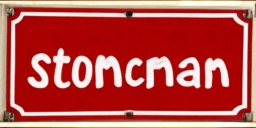}
\includegraphics[width=0.24\columnwidth]{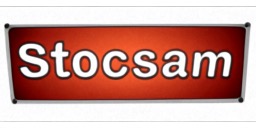}
\includegraphics[width=0.24\columnwidth]{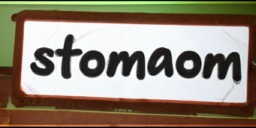}

\vspace{1ex}

ByT5-XXL \\[0.5ex]

\includegraphics[width=0.23\columnwidth, cfbox=white 1pt 0pt]{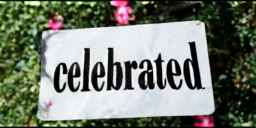}
\includegraphics[width=0.23\columnwidth, cfbox=white 1pt 0pt]{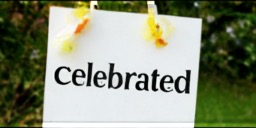}
\includegraphics[width=0.23\columnwidth, cfbox=red 1pt 0pt]{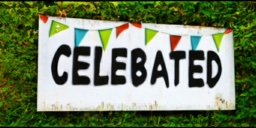}
\includegraphics[width=0.23\columnwidth, cfbox=red 1pt 0pt]{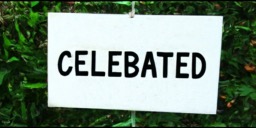}

\includegraphics[width=0.23\columnwidth, cfbox=white 1pt 0pt]{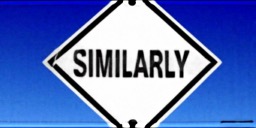}
\includegraphics[width=0.23\columnwidth, cfbox=white 1pt 0pt]{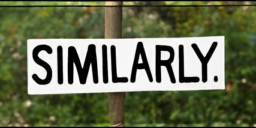}
\includegraphics[width=0.23\columnwidth, cfbox=red 1pt 0pt]{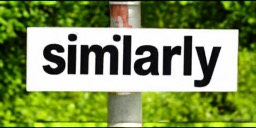}
\includegraphics[width=0.23\columnwidth, cfbox=red 1pt 0pt]{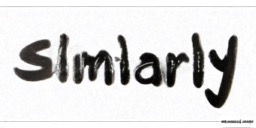}

\includegraphics[width=0.23\columnwidth, cfbox=white 1pt 0pt]{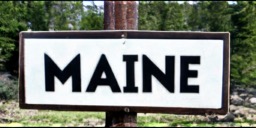}
\includegraphics[width=0.23\columnwidth, cfbox=white 1pt 0pt]{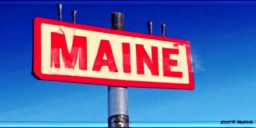}
\includegraphics[width=0.23\columnwidth, cfbox=white 1pt 0pt]{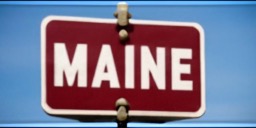}
\includegraphics[width=0.23\columnwidth, cfbox=red 1pt 0pt]{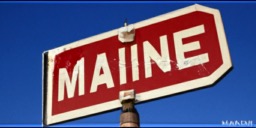}

\caption{Selected errors from two models on DrawText Spelling. \textbf{Top}: Sample words that T5-XXL consistently misspells (target: \textit{refers}, \textit{similarly}, \textit{stomach}). \textbf{Bottom}: ByT5-XXL errors (red outline) are more sporadic and minor: dropped, merged, or repeated glyphs. See Appendix~\ref{sec:appendix_spelling_samples} for more examples.}
\label{fig:spelling-errors}
\end{figure}

\begin{figure}
\centering
\includegraphics[width=\columnwidth, trim={1.5ex, 6.5ex, 2.0ex, 0.0ex}, clip]{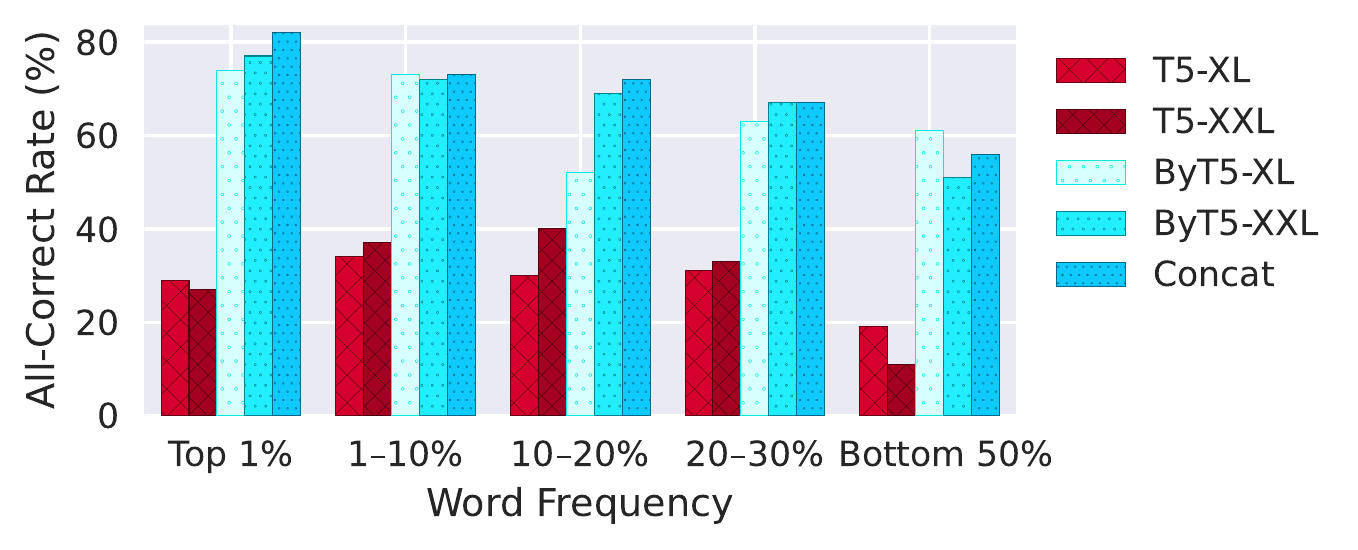}
\includegraphics[width=\columnwidth, trim={1.5ex, 1.5ex, 2.0ex, 0.0ex}, clip]{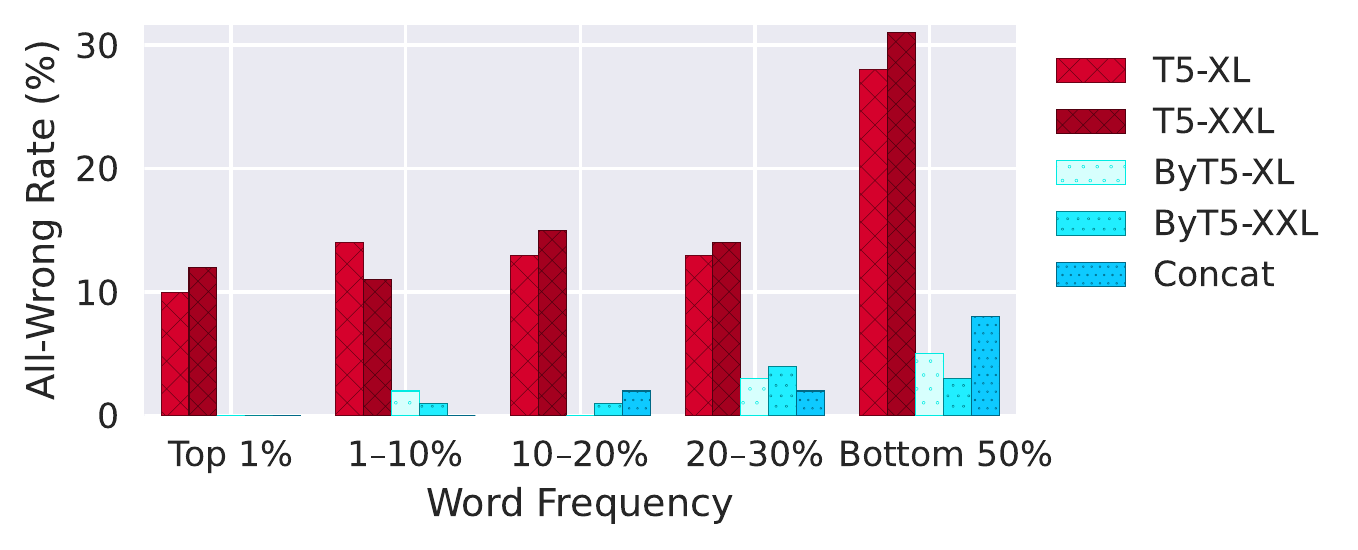}
\caption{Proportion of words for which models are consistently correct (top) or consistently incorrect (bottom) across $4$ image samples on DrawText Spelling.}
\label{fig:ocr_all_right_all_wrong}
\end{figure}

\paragraph{Char-aware models reduce consistent errors.} Another stark difference between our models lies in whether they \emph{consistently} misspell a given word across multiple samples. As illustrated in Figure~\ref{fig:spelling-errors}, there are many words that our T5 models misspell no matter how many samples are drawn. Again, we believe this indicates missing knowledge in the text encoder. By contrast, our ByT5 models are more likely to make sporadic errors. We quantify this observation in Figure~\ref{fig:ocr_all_right_all_wrong} by measuring the rates at which the model is consistently right ($4/4$) or wrong ($0/4$) across all four image samples. On common words in particular (Top~1\%), we see a sharp contrast in that ByT5 models are \emph{never} consistently wrong, while T5 models are consistently wrong on $10$\% or more of words.

\subsection{DrawText Creative results}

To test our models in a more realistic user-facing setting, we sample $8$ images from each of our T5 and ByT5 models on our $175$ DrawText Creative prompts in Appendix~\ref{sec:appendix_creative_prompts}. These prompts are more diverse and challenging, with the majority targeting three or more words of rendered text.

Focusing on text rendering ability,\footnote{We note our models' overall image quality and alignment fall short of a state-of-art model like Imagen (see Figure~\ref{fig:creative_prompts}). This is expected, given that our models train exclusively on the lightly curated Laion-400M dataset \cite{laion400m}, and see $5.6\times$ fewer examples than Imagen during training.} we find once again that character-aware models have a clear advantage. Figures~\ref{fig:exquisite} and \ref{fig:snowmen} show non-cherrypicked samples on two prompts where T5-XXL consistently misspells one or more words. On prompts targeting longer text spans, all our models struggle, as seen in Figure~\ref{fig:chimpanzees}. Nevertheless, we observe that character-aware text encoders provide a clear lift on these prompts, reducing the misspellings of words like \texttt{refrain}, \texttt{arguing}, and \texttt{chimpanzees}.

\begin{figure}
\centering
\includegraphics[width=0.85\columnwidth, trim={1ex, 1.5ex, 1ex, 1ex}, clip]{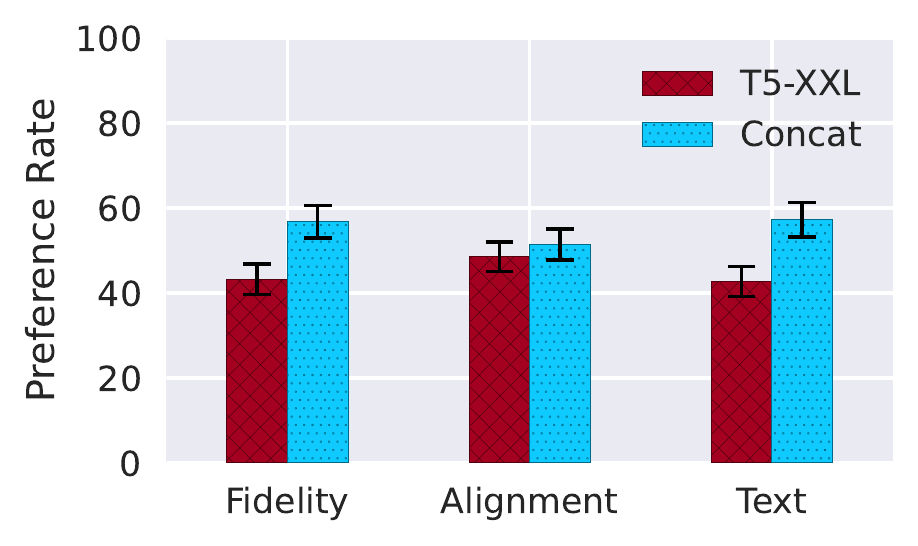}
\caption{On DrawText Creative, our Concat model is preferred over T5-XXL on fidelity and alignment, as well as a new metric assessing rendered text accuracy.}
\label{fig:creative_human_eval}
\end{figure}

We confirm the above observations quantitatively by comparing T5-XXL vs.~Concat using the DrawBench methodology \cite{imagen}, evaluated over our $175$ creative prompts. Beyond the standard DrawBench metrics of fidelity and alignment (described in the following section), we ask raters \textit{Which set of images more accurately shows the text: ``<\texttt{target text}>''?} Results in Figure~\ref{fig:creative_human_eval} show Concat is preferred on all metrics.

\subsection{DrawBench results}

We have shown that character-aware text encoders excel at spelling, in both text (\S\ref{sec:text}) and visual (\S\ref{sec:image_experiments}) domains. But does this ability come at a cost? Can these models maintain competitive image quality and text-image alignment, even on prompts that don't require text rendering? To shed light on this question, we run several side-by-side comparisons using the DrawBench evaluation of \citet{imagen}. This asks human raters to compare two models' generations of $8$ images each across $200$ prompts covering $11$ thematic categories. We follow the procedure described in \citet{imagen} closely, aggregating scores across $25$ raters.

\begin{figure}
\centering
\includegraphics[height=0.176\textwidth, trim={1.5ex, 1.4ex, 1.5ex, 0}, clip]{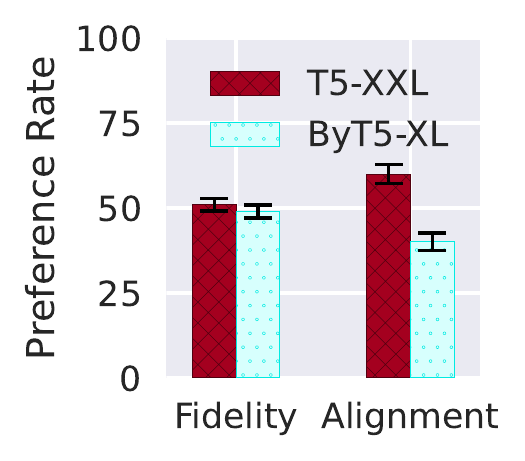} \includegraphics[height=0.176\textwidth, trim={9ex, 1.4ex, 1.5ex, 0}, clip]{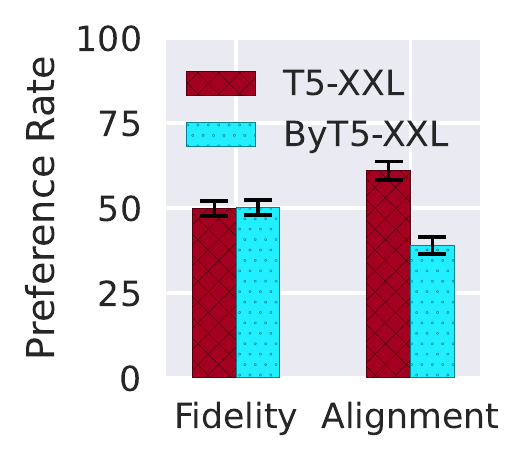}
\includegraphics[height=0.176\textwidth, trim={9ex, 1.4ex, 1.5ex, 0}, clip]{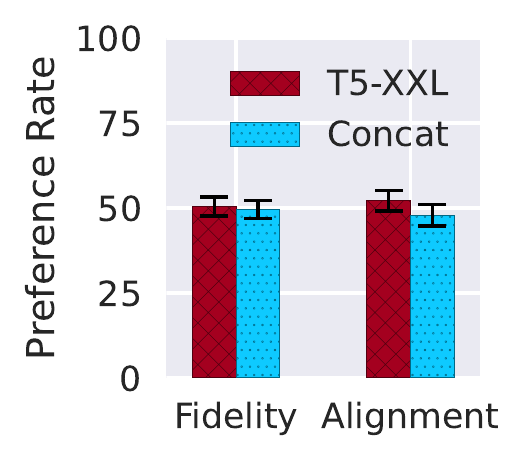}
\caption{DrawBench user preference rates comparing T5-XXL with three character-aware models. While image fidelity is comparable, the pure ByT5 models have lower image-text alignment. Concat closes the alignment gap. Error bars show $95$\% confidence intervals.}
\label{fig:drawbench}
\end{figure}

Figure~\ref{fig:drawbench} shows DrawBench results of three side-by-side comparisons of character-aware models vs.~T5-XXL\@. While image quality (``fidelity'') is similar across the board, we find that \emph{purely} character-level models (ByT5) score worse on image-text alignment, with raters preferring T5-XXL on $60$\% of prompts. By contrast, our Concat model closes this gap to within error bars. Thus, this ``hybrid'' character-aware model is able to greatly improve text rendering (Figure~\ref{fig:ocr_accuracy}), without significantly hurting performance elsewhere.

Appendix~\ref{sec:appendix_drawbench} provides a per-category breakdown and further analysis; while the character-aware models excel on the DrawBench \texttt{text} category, the ByT5 models are dispreferred in most other categories. Through manual inspection, we find the ByT5 models are more prone to ignore information in the prompt, for example leaving out a mentioned object, or choosing a canonical color over a requested one. One possible explanation for this behavior is that we did not tune the \textit{guidance weight} parameter used at inference time \cite{imagen}, using a fixed value of $30$ throughout. Increasing this parameter is known to boost image-text alignment, but at the cost of diversity. It may be that character-level models benefit from higher guidance values than token-based models.

Another possibility is that the ByT5 models have a shallower understanding of English language due to their multilingual nature---as ByT5 was exposed to roughly $70\times$ less English than T5 during pretraining.\footnote{The models were trained on the same number of tokens, but only $6$\% of ByT5 training was on English, and we estimate $4$ UTF-8 bytes per T5 token.}
Given this difference, we should also expect to see corresponding gains on non-English languages. We confirm this expectation through preliminary results in Appendix~\ref{sec:appendix_multilingual}.

\section{Conclusion}

In this paper, we set out to better understand what is needed for image generation models to reliably render well-formed visual text. Using our novel WikiSpell and DrawText benchmarks, we were able to precisely quantify the effects of character-awareness and other design choices on spelling ability in both the text and visual domains.

We found that character-aware text encoders confer large gains on spelling, and when used within an image generation model, these gains translate into improved visual text rendering. However, using \emph{exclusively} character-level representations deteriorated overall text-image alignment---at least when evaluating our multilingual ByT5 text encoder on English prompts with untuned guidance weight. To resolve this, we found that a hybrid model combining token-level and character-level signals offered the best of both worlds: dramatically improving visual text without significantly affecting alignment.

While we saw substantial improvements on DrawText Spelling accuracy ($75$\%\,$\rightarrow$\,$94$\% on common words and $47$\%\,$\rightarrow$\,$83$\% on rare words), some failure modes remain unaddressed. Even our strongest models were observed to occasionally drop, repeat, or merge letters within a word, or words within a phrase. Our results strongly suggest that resolving these issues will require orthogonal improvements outside the text encoder, specifically changes to the image generation module.

As a secondary finding, we demonstrated for the first time that, with sufficient scale, even models lacking a direct character-level view of their inputs can \emph{infer} robust spelling information through knowledge gained via web pretraining---``the spelling miracle''. While remarkable, this finding is less immediately practical, as it requires models over 100B parameters, and even these did not generalize well beyond English in our experiments.

\section*{Limitations}

While we establish the ``miraculous'' ability of character-blind models to induce robust spelling information through large-scale web pretraining, our work does not attempt to identify the mechanisms or sources through which this information is learned. Possible sources within web corpora include: dictionaries containing phonetic pronunciation guides, alphabetically ordered lists, typos and other misspellings, and examples of spelling words with dashes or spaces between every character. Linguistic phenomena that may aide in inducing spelling knowledge include words with predictable morphemic makeup, and cases where meaning-form relation is non-arbitrary, contra Saussure’s ``semiotic arbitrariness''. We refer the reader to \citet{itzhak-levy-2022-models} and \citet{kaushal-mahowald-2022-tokens} for work in this direction.

Most of our image generation experiments are limited to English. We present preliminary results in Appendix~\ref{sec:appendix_multilingual} showing that our ByT5-based models have stronger multilingual understanding than T5. However it would be valuable to test this further, and to explore training image generation models on multilingual image-caption datasets, as opposed to merely using a pretrained multilingual text encoder.

Ideally, it would be possible to conduct controlled comparisons between pretrained text encoders that differ only in one regard, to isolate all factors contributing to performance. However as pretraining large language models is resource intensive, we were only able to use off-the-shelf text encoders, which often differ along multiple axes. In our text-only experiments, we isolated the contributions of character-awareness (ByT5 vs.~mT5/T5) and multilinguality (ByT5/mT5 vs.~T5). However, in our image generation experiments, these factors were conflated, as we had limited resources for training new models. Still, the fact that ByT5-based image generation models outperform T5 \emph{despite} being multilingual (which often degrades performance on English-only tasks) strongly suggests that character-awareness is the key factor for spelling ability.

Another limitation is that we focused on image generation models that leverage \emph{frozen} pretrained text encoders. This enabled straightforward experimentation by swapping encoders and retraining the image generation module. However, it remains to be seen whether our results extend to settings where the text encoder is trained along with the rest of the model, as in \citet{parti}.

\section*{Ethics Statement}

We note our image-caption training data comes from Laion-400M \cite{laion400m}, which is uncurated and known to contain harmful biases and offensive content. We hope to utilize and contribute to safer and better curated datasets in the future, as well as to develop improved techniques for debiasing and detoxifying existing models.

A potential risk for image generation models is that they can be used for creating misleading and harmful content. Our work on improving text rendering could aide the creation of fake signs and other misleading images containing visual text. With the wide-spread availability of image generation models, we expect that improving education around misinformation and adopting better digital signature mechanisms will be important countermeasures.

\section*{Acknowledgements}

We thank Jason Baldridge, Jon Clark, Noah Fiedel, Linting Xue, and Jiahui Yu for helpful discussion and comments on an earlier draft. We thank Sarah Pratt for validating findings on Stable Diffusion models.

\bibliography{anthology,custom}
\bibliographystyle{acl_natbib}

\appendix

\section{Multilingual results}
\label{sec:appendix_multilingual}

\begin{figure*}
\centering

T5-XXL \\[0.3ex]
\includegraphics[width=0.073\textwidth, cfbox=white 1pt 0pt]{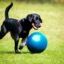}
\includegraphics[width=0.073\textwidth, cfbox=white 1pt 0pt]{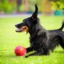}
\includegraphics[width=0.073\textwidth, cfbox=white 1pt 0pt]{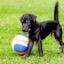}
\includegraphics[width=0.073\textwidth, cfbox=white 1pt 0pt]{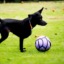}
\includegraphics[width=0.073\textwidth, cfbox=red 1pt 0pt]{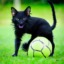}
\includegraphics[width=0.073\textwidth, cfbox=red 1pt 0pt]{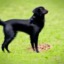}
\includegraphics[width=0.073\textwidth, cfbox=red 1pt 0pt]{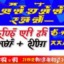}
\includegraphics[width=0.073\textwidth, cfbox=red 1pt 0pt]{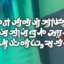}
\includegraphics[width=0.073\textwidth, cfbox=red 1pt 0pt]{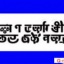}
\includegraphics[width=0.073\textwidth, cfbox=red 1pt 0pt]{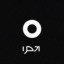}
\includegraphics[width=0.073\textwidth, cfbox=red 1pt 0pt]{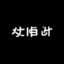}
\includegraphics[width=0.073\textwidth, cfbox=red 1pt 0pt]{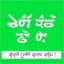}

\includegraphics[width=0.073\textwidth, cfbox=white 1pt 0pt]{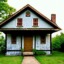}
\includegraphics[width=0.073\textwidth, cfbox=white 1pt 0pt]{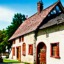}
\includegraphics[width=0.073\textwidth, cfbox=white 1pt 0pt]{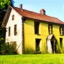}
\includegraphics[width=0.073\textwidth, cfbox=white 1pt 0pt]{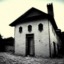}
\includegraphics[width=0.073\textwidth, cfbox=white 1pt 0pt]{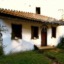}
\includegraphics[width=0.073\textwidth, cfbox=white 1pt 0pt]{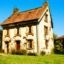}
\includegraphics[width=0.073\textwidth, cfbox=red 1pt 0pt]{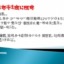}
\includegraphics[width=0.073\textwidth, cfbox=red 1pt 0pt]{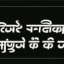}
\includegraphics[width=0.073\textwidth, cfbox=red 1pt 0pt]{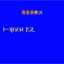}
\includegraphics[width=0.073\textwidth, cfbox=red 1pt 0pt]{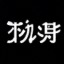}
\includegraphics[width=0.073\textwidth, cfbox=red 1pt 0pt]{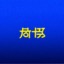}
\includegraphics[width=0.073\textwidth, cfbox=red 1pt 0pt]{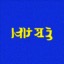}

\includegraphics[width=0.073\textwidth, cfbox=white 1pt 0pt]{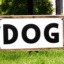}
\includegraphics[width=0.073\textwidth, cfbox=white 1pt 0pt]{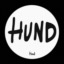}
\includegraphics[width=0.073\textwidth, cfbox=red 1pt 0pt]{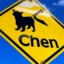}
\includegraphics[width=0.073\textwidth, cfbox=white 1pt 0pt]{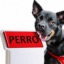}
\includegraphics[width=0.073\textwidth, cfbox=white 1pt 0pt]{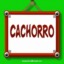}
\includegraphics[width=0.073\textwidth, cfbox=red 1pt 0pt]{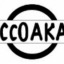}
\includegraphics[width=0.073\textwidth, cfbox=red 1pt 0pt]{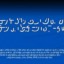}
\includegraphics[width=0.073\textwidth, cfbox=red 1pt 0pt]{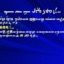}
\includegraphics[width=0.073\textwidth, cfbox=red 1pt 0pt]{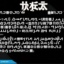}
\includegraphics[width=0.073\textwidth, cfbox=red 1pt 0pt]{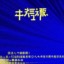}
\includegraphics[width=0.073\textwidth, cfbox=red 1pt 0pt]{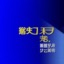}
\includegraphics[width=0.073\textwidth, cfbox=red 1pt 0pt]{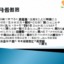}

\vspace{0.5ex}

ByT5-XXL \\

\includegraphics[width=0.073\textwidth, cfbox=white 1pt 0pt]{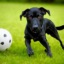}
\includegraphics[width=0.073\textwidth, cfbox=white 1pt 0pt]{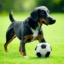}
\includegraphics[width=0.073\textwidth, cfbox=white 1pt 0pt]{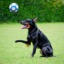}
\includegraphics[width=0.073\textwidth, cfbox=white 1pt 0pt]{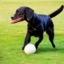}
\includegraphics[width=0.073\textwidth, cfbox=white 1pt 0pt]{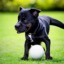}
\includegraphics[width=0.073\textwidth, cfbox=white 1pt 0pt]{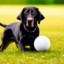}
\includegraphics[width=0.073\textwidth, cfbox=white 1pt 0pt]{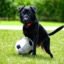}
\includegraphics[width=0.073\textwidth, cfbox=white 1pt 0pt]{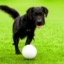}
\includegraphics[width=0.073\textwidth, cfbox=white 1pt 0pt]{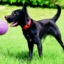}
\includegraphics[width=0.073\textwidth, cfbox=white 1pt 0pt]{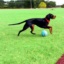}
\includegraphics[width=0.073\textwidth, cfbox=white 1pt 0pt]{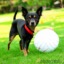}
\includegraphics[width=0.073\textwidth, cfbox=white 1pt 0pt]{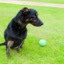}

\includegraphics[width=0.073\textwidth, cfbox=white 1pt 0pt]{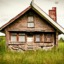}
\includegraphics[width=0.073\textwidth, cfbox=white 1pt 0pt]{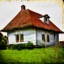}
\includegraphics[width=0.073\textwidth, cfbox=white 1pt 0pt]{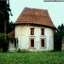}
\includegraphics[width=0.073\textwidth, cfbox=white 1pt 0pt]{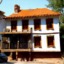}
\includegraphics[width=0.073\textwidth, cfbox=white 1pt 0pt]{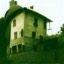}
\includegraphics[width=0.073\textwidth, cfbox=white 1pt 0pt]{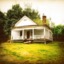}
\includegraphics[width=0.073\textwidth, cfbox=white 1pt 0pt]{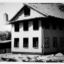}
\includegraphics[width=0.073\textwidth, cfbox=white 1pt 0pt]{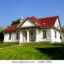}
\includegraphics[width=0.073\textwidth, cfbox=white 1pt 0pt]{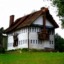}
\includegraphics[width=0.073\textwidth, cfbox=white 1pt 0pt]{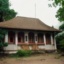}
\includegraphics[width=0.073\textwidth, cfbox=white 1pt 0pt]{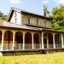}
\includegraphics[width=0.073\textwidth, cfbox=white 1pt 0pt]{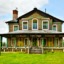}

\includegraphics[width=0.073\textwidth, cfbox=white 1pt 0pt]{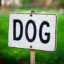}
\includegraphics[width=0.073\textwidth, cfbox=white 1pt 0pt]{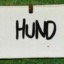}
\includegraphics[width=0.073\textwidth, cfbox=white 1pt 0pt]{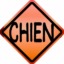}
\includegraphics[width=0.073\textwidth, cfbox=white 1pt 0pt]{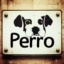}
\includegraphics[width=0.073\textwidth, cfbox=white 1pt 0pt]{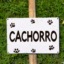}
\includegraphics[width=0.073\textwidth, cfbox=red 1pt 0pt]{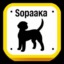}
\includegraphics[width=0.073\textwidth, cfbox=red 1pt 0pt]{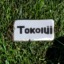}
\includegraphics[width=0.073\textwidth, cfbox=red 1pt 0pt]{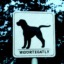}
\includegraphics[width=0.073\textwidth, cfbox=red 1pt 0pt]{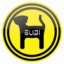}
\includegraphics[width=0.073\textwidth, cfbox=red 1pt 0pt]{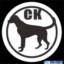}
\includegraphics[width=0.073\textwidth, cfbox=red 1pt 0pt]{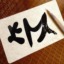}
\includegraphics[width=0.073\textwidth, cfbox=red 1pt 0pt]{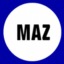}

\texttt{en} \hspace{1.96em} \texttt{de} \hspace{1.96em} \texttt{fr} \hspace{1.96em} \texttt{es} \hspace{1.96em} \texttt{pt} \hspace{1.96em} \texttt{ru} \hspace{1.96em} \texttt{el} \hspace{1.96em} \texttt{hi} \hspace{1.96em} \texttt{ar} \hspace{1.96em} \texttt{zh} \hspace{1.96em} \texttt{ja} \hspace{1.96em} \texttt{ko}

\caption{Our T5-XXL model (top) shows some multilingual ability in high-resource European languages (German, French, Spanish, Portuguese, Russian), but fails on Greek, Hindi, Arabic, Chinese, Japanese, and Korean. Our ByT5-XXL model (bottom) exhibits understanding across all of these languages. However, neither model is capable of accurately rendering text in non-Latin scripts (last $7$ columns of rows $3$ and $6$). Prompts are translations via Google Translate of: (1)~\textit{A photo of a black dog playing with a ball.} (2)~\textit{A photo of an old house.} (3)~\textit{A sign with the word "dog" written on it.} Inaccurate renderings are outlined in red.}
\label{fig:multilingual}
\end{figure*}

As ByT5 is a multilingual model covering $100$+ languages, we are interested to see if image generation models built on ByT5 deliver improved performance over T5 on non-English languages. While the text encoder itself is multilingual, it is not obvious whether this is sufficient to produce a multilingual image generation model. The image caption dataset used for training in all of our experiments is Laion-400M \cite{laion400m}, which we estimate through language ID detection to consist of $95$\% English captions, with only minimal coverage (<$0.1$\%) of some widely spoken languages, such as Arabic and Hindi.

To test for multilingual understanding, we translate two English prompts to $11$ languages using Google Translate, and feed the outputs to our models. As can be seen in the rows $1$--$2$ of Figure~\ref{fig:multilingual}, our T5-XXL model demonstrates basic understanding of five high-resource European languages (German, French, Spanish, Portuguese, Russian).\footnote{A few minor problems are visible: swapping \textit{dog} $\rightarrow$ \textit{cat} in Portuguese, and not rendering a ball in Russian.} However, in the remaining languages (Greek, Hindi, Arabic, Chinese, Japanese, Korean), T5 appears to ignore the caption completely.

By comparison, our ByT5-XXL model exhibits understanding across all $11$ languages. Given its limited training on multilingual captions, we interpret this ability as due to the pretrained ByT5 encoder's alignment of representations across languages. If the encoder already embeds similar prompts into a shared space that factors out the contribution of language, then the image generation model should be able to learn from just a handful of examples how to map any language seen in pretraining into the space of images.\footnote{We observe in several examples that the prompt language can bias the model towards culturally-relevant visual interpretations. For example, the Chinese prompt for \textit{A photo of an old house} (\chinese{一张老房子的照片}) produces a house with a curved roof. It would be interesting to further explore the extent of these biases and the degree to which they can be overcome where unwanted.}

If this explanation is correct, it also suggests that rendering text in different scripts will require more than just a multilingual encoder. To learn the glyph shapes, variants and fonts used for a given script, we should expect to need to train models on a large source of visual text in that script. Indeed, rows $3$ and $6$ of Figure~\ref{fig:multilingual} show that neither of our models can map prompt text onto visual text in non-Latin scripts. While our ByT5 model captures the intent to draw a sign across all languages, it is unable to render the words for \textit{dog} in Greek, Russian, Chinese and so on, presumably because it has had little visual exposure to the glyphs making up these words.\footnote{Interestingly, in Russian, the model is able to nearly-successfully transliterate \russian{собака} (dog) to Latin script, as \texttt{sopaaka}. We suspect this transliteration ability is learned during the text encoder pretraining \citep{pires-etal-2019-multilingual}.}

\section{OCR error estimation}
\label{sec:appendix_ocr_validate}

To estimate the rate of false positives and false negatives due to OCR errors, we sample $32$ examples labeled \textit{correct} and $32$ labeled \textit{incorrect} for each of T5-XXL and ByT5-XXL, and perform a manual validation. In our sample, we find no false positives: when OCR detects the correct word, it is always correct. However observe false negatives for both models, including cases where OCR fails to detect the text (e.g., due to it being too small), or misreads a character. For ByT5-XXL, we find that $34$\% of examples labeled by OCR as \textit{incorrect} are actually correct. For T5-XXL, this error rate is lower at $9$\%. This asymmetry suggests that the benefit of character-aware modeling may be even greater than implied by our results in Figure~\ref{fig:ocr_accuracy}.

\section{Per-category DrawBench analysis}
\label{sec:appendix_drawbench}

\begin{figure}
\centering
\includegraphics[width=\columnwidth, trim={0.8ex, 13.1ex, 0.8ex, 0}, clip]{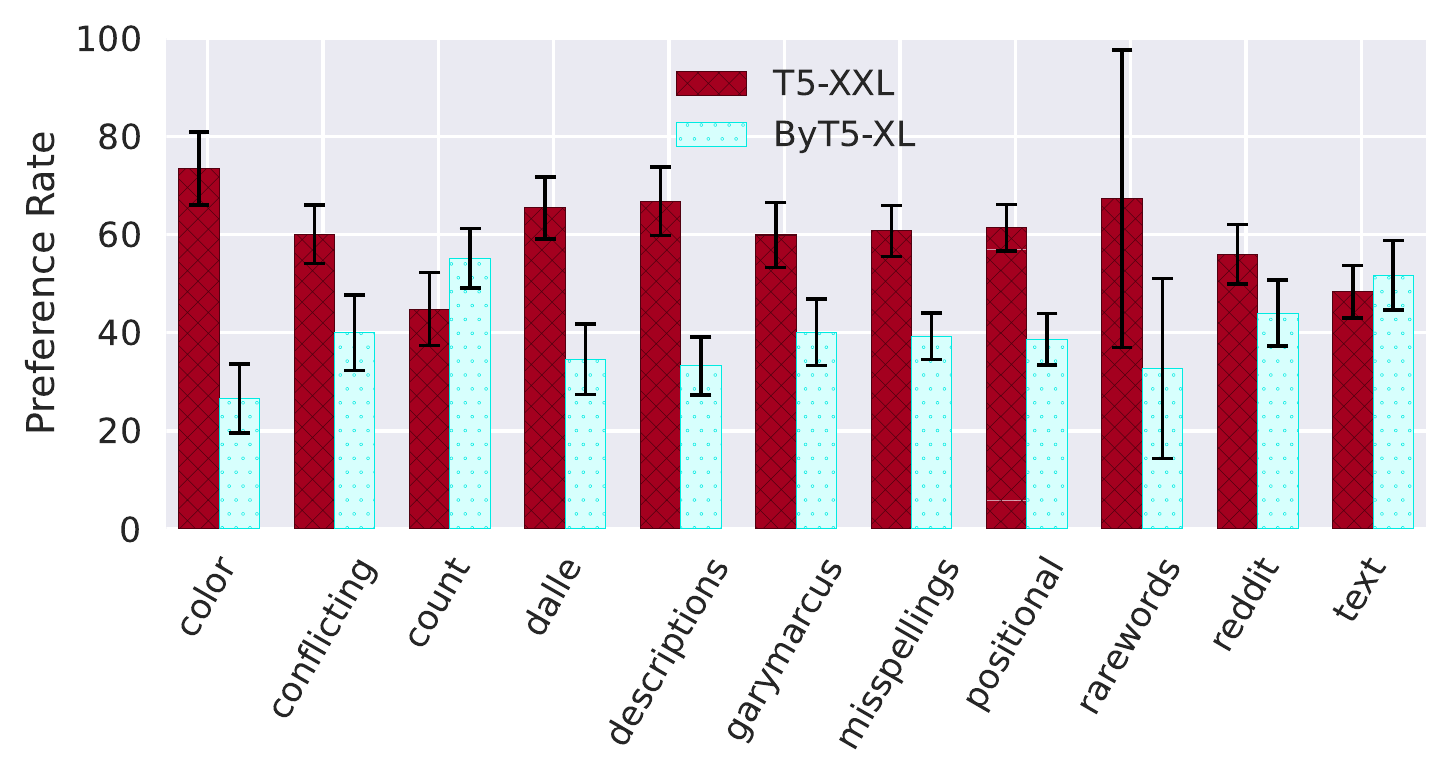}
\includegraphics[width=\columnwidth, trim={0.8ex, 13.1ex, 0.8ex, 0}, clip]{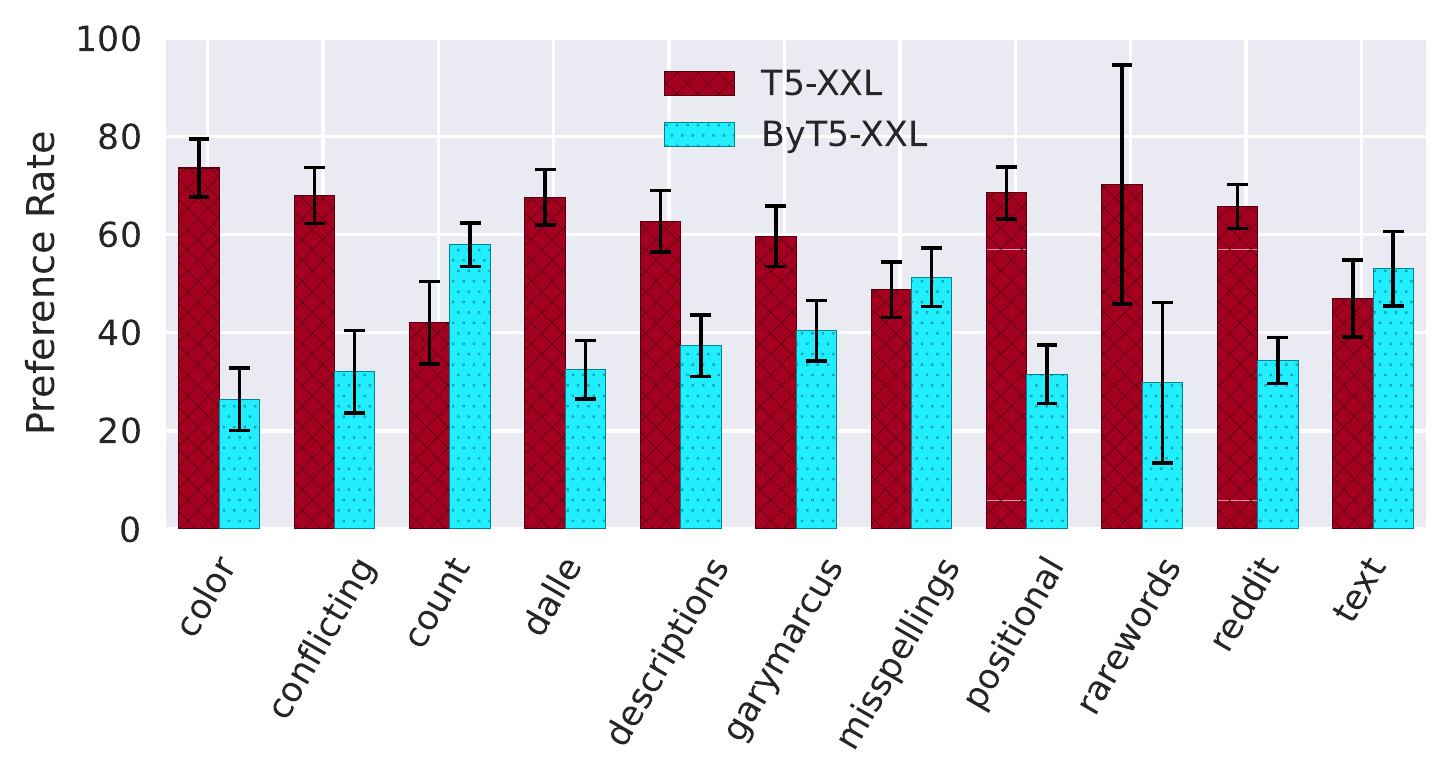}
\includegraphics[width=\columnwidth, trim={0.8ex, 1.9ex, 0.8ex, 0}, clip]{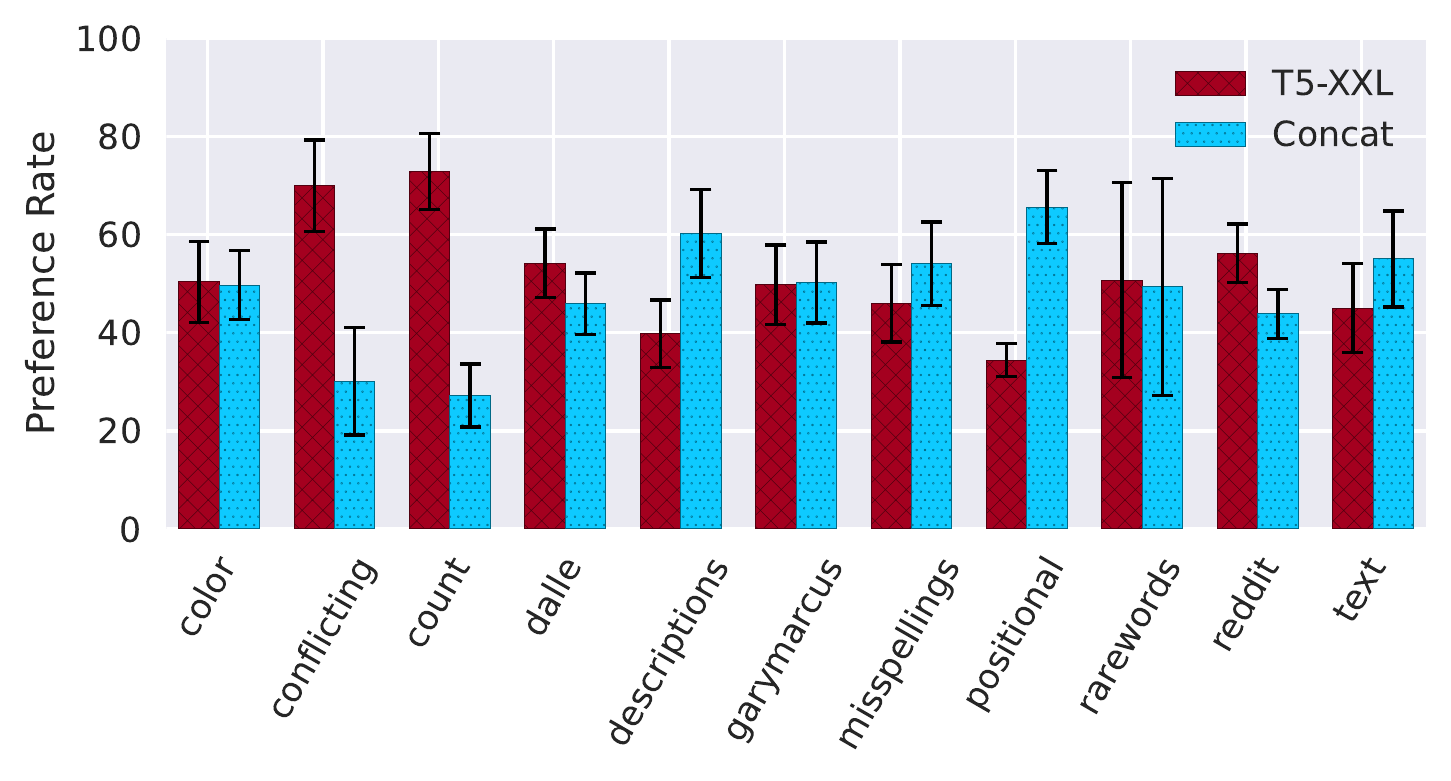}
\caption{Per-category DrawBench image-text alignment preferences comparing T5-XXL with three character-aware models. All character-aware models are preferred on the \texttt{text} category. On other categories, while pure ByT5 models are generally dispreferred, the Concat model is competitive with T5-XXL.}
\label{fig:drawbench_alignment}
\end{figure}

To understand the alignment scores in more detail, we report per-category preference scores in Figure~\ref{fig:drawbench_alignment}. In line with our DrawText Spelling results, the character-aware models are always preferred in the \texttt{text} category---$21$ prompts testing the ability to render $7$ short phrases in $3$ visual styles. The ByT5 models are also preferred in the \texttt{count} category, which tests prompts like \textit{Four dogs on the street}. However, they are dispreferred in nearly all other cases, and perform particularly poorly on the \texttt{color} category. Through manual inspection, we find that in this category, the ByT5 models are more prone to ignore information in the prompt, for example leaving out a mentioned object, or choosing a canonical color over a requested one (e.g.~a yellow banana instead of a red one).

\section{WikiSpell details}
\label{sec:appendix_wikispell}

We select six languages to cover diverse propoerties that could affect the ability for models to learn spellings: \textbf{Arabic}, written in the Arabic alphabet, has non-concatenative morphology;
\textbf{Chinese} is written in Simplified and Traditional Chinese scripts, which are logographic and do not use whitespace to separate words;
\textbf{Finnish}, written in the Latin alphabet, has rich inflectional and derivational suffixes, and word stems often change when suffixes are attached;
\textbf{Korean}'s writing system, Hangul, has a huge number of characters since alphabetic features are arranged into syllabic blocks, which Unicode represents as a single characters;
\textbf{Russian}, written in the Cyrillic alphabet, has substantial fusional morphology, and uses inflection for case-marking and agreement; and
\textbf{Thai}, written in the alphabetic Thai script, is an analytic language, but does not use whitespace between words.

Further implementation details are as follows:

\begin{itemize}
\setlength{\itemsep}{4pt}
\setlength{\itemindent}{0pt}
\setlength{\parskip}{0pt}
\setlength{\parsep}{0pt}
\item Example Python 3 code for transforming a word into its spelling:
\begin{verbatim}
  def to_spelling(word: str) -> str:
    return " ".join(word)
\end{verbatim}
\item Since we want each entry to be a single word, we exclude entries that contain any (Unicode) whitespace, that are entirely punctuation/symbols (i.e., all characters are from Unicode categories P and/or S), that are longer than $30$ characters, or that have a ``part-of-speech'' \textit{Proverb}.
\item For efficiency, word frequencies are computed on \emph{subsets} of the full mC4 corpus. 
For languages other than English, this is a sample of 1M documents from that language's section of mC4.
For English, since it has such a long tail of words in Wiktionary, we use the first 140M documents in mC4's English section.
\item For Arabic, English, Finnish, Korean, and Russian, word-counting is performed by splitting document texts using the following delimiters: \texttt{?!/:;,\textbackslash{}"\&()[]\{\}<>\`}, plus any Unicode whitespace. For Chinese and Thai, since they do not use whitespace to separate words, we instead count the number of documents in which the word appeared as a substring.
\end{itemize}

\section{Additional DrawText creative samples}
\label{sec:appendix_creative_samples}

\begin{figure*}
\centering

T5-XXL \\[0.5ex]
\includegraphics[width=0.24\textwidth]{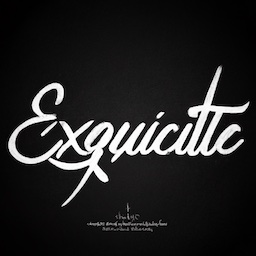}
\includegraphics[width=0.24\textwidth]{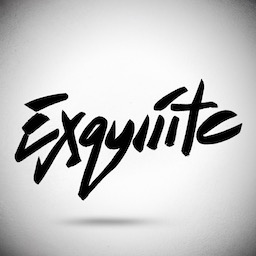}
\includegraphics[width=0.24\textwidth]{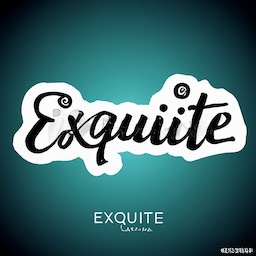}
\includegraphics[width=0.24\textwidth]{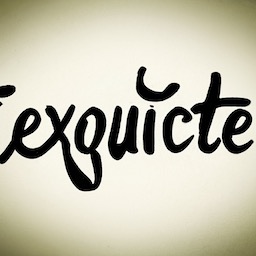}
\includegraphics[width=0.24\textwidth]{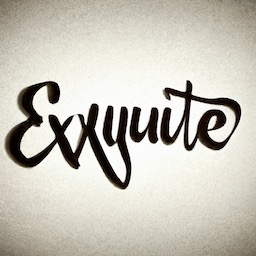}
\includegraphics[width=0.24\textwidth]{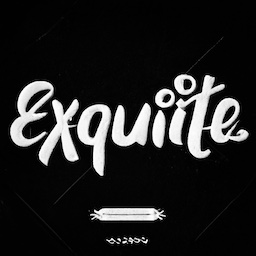}
\includegraphics[width=0.24\textwidth]{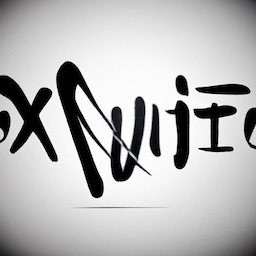}
\includegraphics[width=0.24\textwidth]{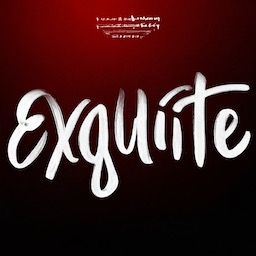}

\vspace{0.5ex}

ByT5-XXL \\[0.3ex]
\includegraphics[width=0.24\textwidth]{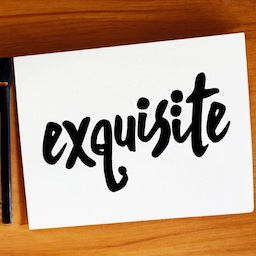}
\includegraphics[width=0.24\textwidth]{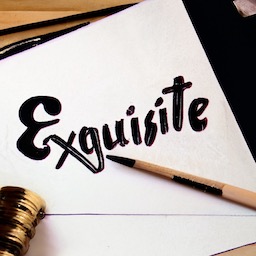}
\includegraphics[width=0.24\textwidth]{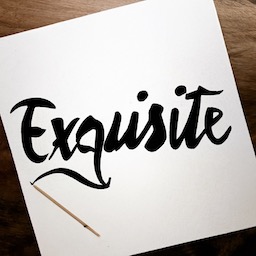}
\includegraphics[width=0.24\textwidth]{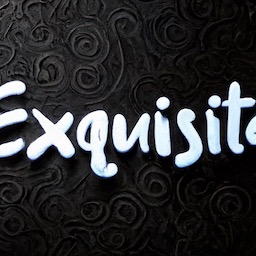}
\includegraphics[width=0.24\textwidth]{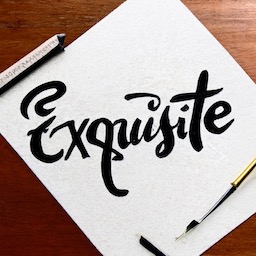}
\includegraphics[width=0.24\textwidth]{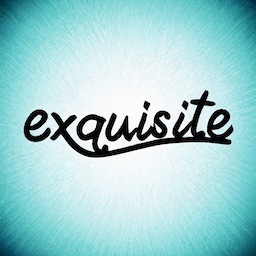}
\includegraphics[width=0.24\textwidth]{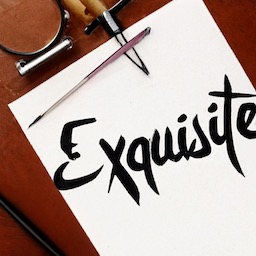}
\includegraphics[width=0.24\textwidth]{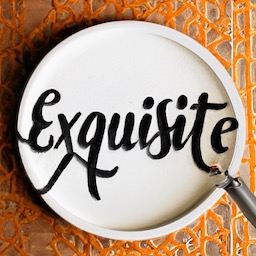}

\caption{Non-cherrypicked samples from our T5-XXL (top) and ByT5-XXL (bottom) models. The character-aware ByT5 model reliably spells the target word correctly, with only minor issues around letter shapes or letter merging. Over 100 samples, we found the character-blind T5 model never produced the target spelling. Prompt: \textit{The word "exquisite" written in modern calligraphy.}}
\label{fig:exquisite}
\end{figure*}

\begin{figure*}
\centering

T5-XXL \\[0.5ex]
\includegraphics[width=0.24\textwidth]{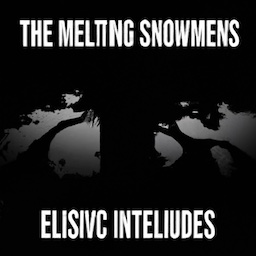}
\includegraphics[width=0.24\textwidth]{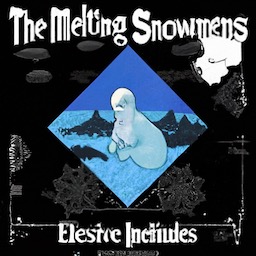}
\includegraphics[width=0.24\textwidth]{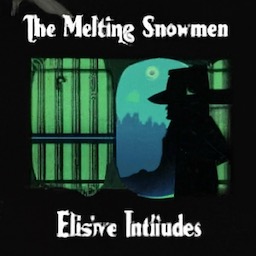}
\includegraphics[width=0.24\textwidth]{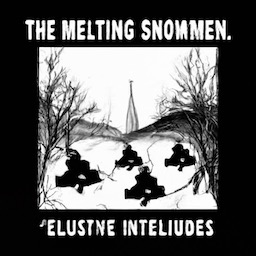}
\includegraphics[width=0.24\textwidth]{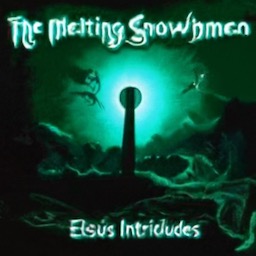}
\includegraphics[width=0.24\textwidth]{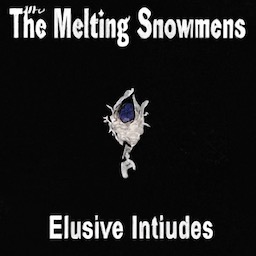}
\includegraphics[width=0.24\textwidth]{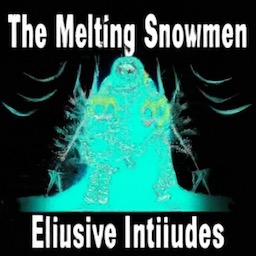}
\includegraphics[width=0.24\textwidth]{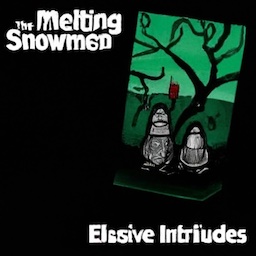}

\vspace{0.5ex}

ByT5-XXL \\[0.3ex]
\includegraphics[width=0.24\textwidth]{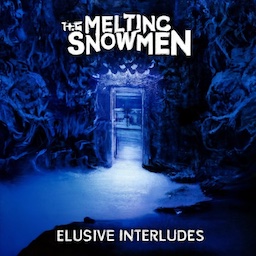}
\includegraphics[width=0.24\textwidth]{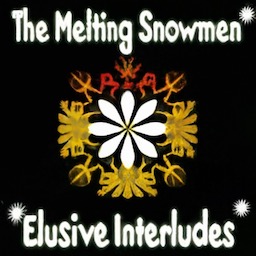}
\includegraphics[width=0.24\textwidth]{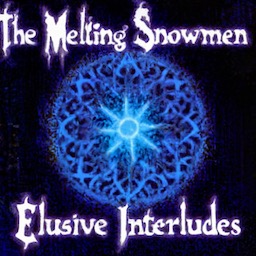}
\includegraphics[width=0.24\textwidth]{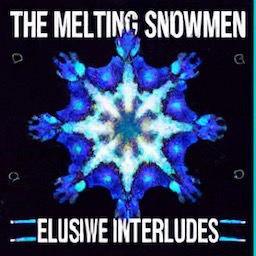}
\includegraphics[width=0.24\textwidth]{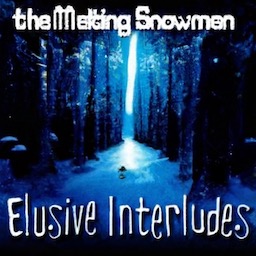}
\includegraphics[width=0.24\textwidth]{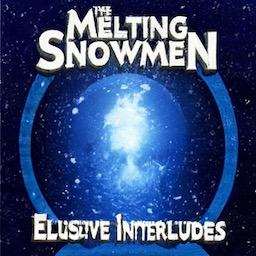}
\includegraphics[width=0.24\textwidth]{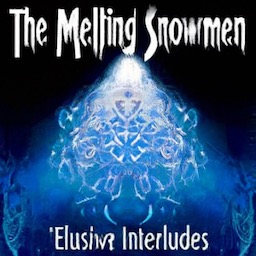}
\includegraphics[width=0.24\textwidth]{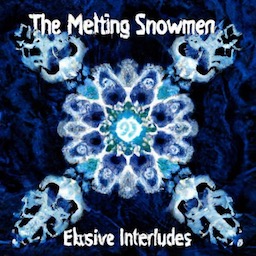}

\caption{Non-cherrypicked samples from our T5-XXL (top) and ByT5-XXL (bottom) models. The character-blind T5 model makes more frequent and more severe errors, including often hallucinating an \texttt{s} at the end of the irregular plural \texttt{snowmen}. Prompt: \textit{The cover for the album 'Elusive Interludes' by the band The Melting Snowmen.} We filter images with no legible text for better comparison, removing a small minority of samples for both models.}
\label{fig:snowmen}
\end{figure*}

\begin{figure*}
\centering

T5-XXL \\[0.5ex]
\includegraphics[width=0.24\textwidth]{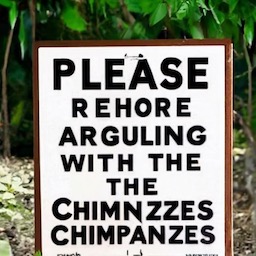}
\includegraphics[width=0.24\textwidth]{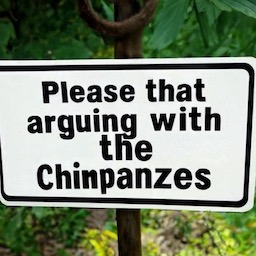}
\includegraphics[width=0.24\textwidth]{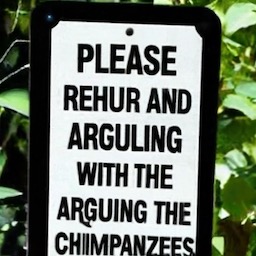}
\includegraphics[width=0.24\textwidth]{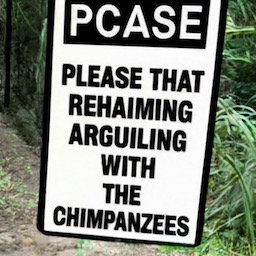}
\includegraphics[width=0.24\textwidth]{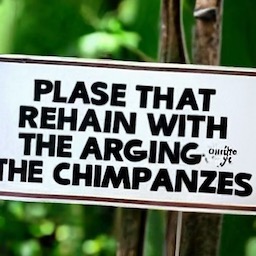}
\includegraphics[width=0.24\textwidth]{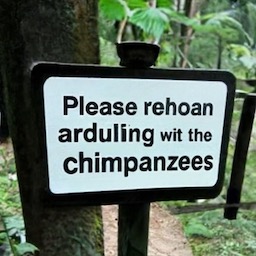}
\includegraphics[width=0.24\textwidth]{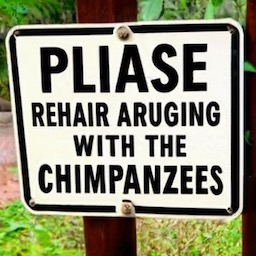}
\includegraphics[width=0.24\textwidth]{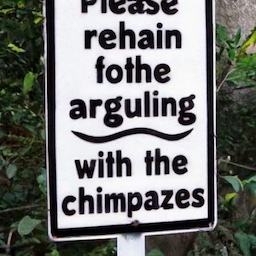}

\vspace{0.5ex}

ByT5-XXL \\[0.3ex]
\includegraphics[width=0.24\textwidth]{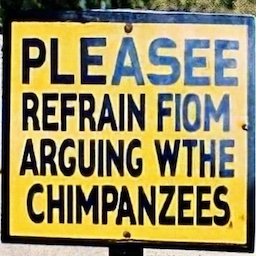}
\includegraphics[width=0.24\textwidth]{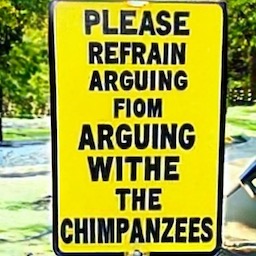}
\includegraphics[width=0.24\textwidth]{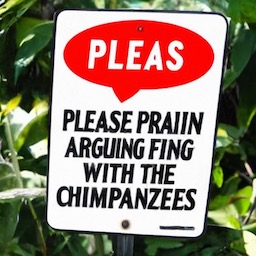}
\includegraphics[width=0.24\textwidth]{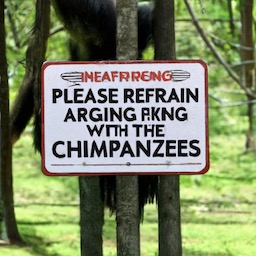}
\includegraphics[width=0.24\textwidth]{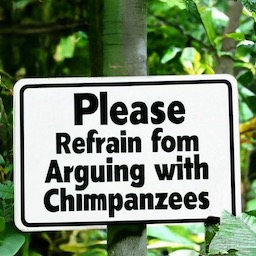}
\includegraphics[width=0.24\textwidth]{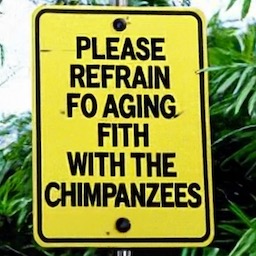}
\includegraphics[width=0.24\textwidth]{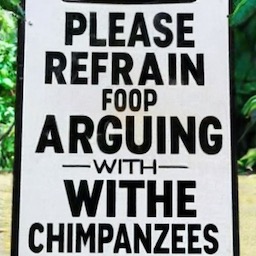}
\includegraphics[width=0.24\textwidth]{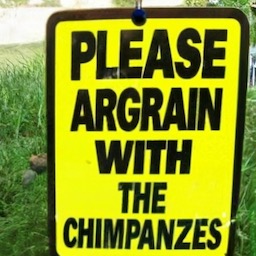}

\caption{Non-cherrypicked samples from our T5-XXL (top) and ByT5-XXL (bottom) models. Both models exhibit layout errors, including dropped/repeated/merged glyphs and words. The T5 model suffers additionally from a lack of core spelling knowledge---misspelling \texttt{refrain}, \texttt{arguing} and \texttt{chimpanzees} on the majority of uses. The ByT5 model is able to spell each of these words correctly in most cases. Prompt: \textit{A sign that says "Please refrain from arguing with the chimpanzees".}}
\label{fig:chimpanzees}
\end{figure*}

We show additional samples on DrawText Creative prompts in Figures \ref{fig:exquisite}, \ref{fig:snowmen} and \ref{fig:chimpanzees}.

\section{Representative DrawText Spelling samples}
\label{sec:appendix_spelling_samples}
\begin{figure*}
\centering
\textit{"0-6-2"} spelled by T5-XL \\[0.5ex]
\includegraphics[width=0.36\columnwidth, cfbox=red 1pt 0pt]{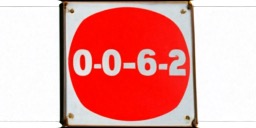}
\includegraphics[width=0.36\columnwidth, cfbox=red 1pt 0pt]{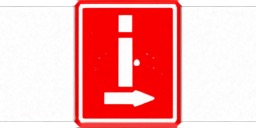}
\includegraphics[width=0.36\columnwidth, cfbox=red 1pt 0pt]{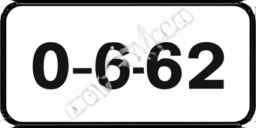}
\includegraphics[width=0.36\columnwidth, cfbox=red 1pt 0pt]{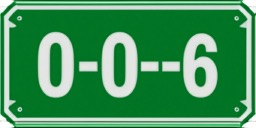}

\vspace{0.5ex}
\textit{"0-6-2"} spelled by T5-XXL \\[0.3ex]
\includegraphics[width=0.36\columnwidth, cfbox=red 1pt 0pt]{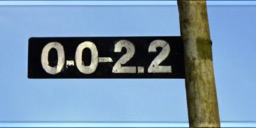}
\includegraphics[width=0.36\columnwidth, cfbox=red 1pt 0pt]{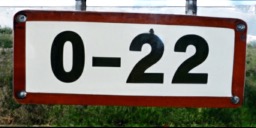}
\includegraphics[width=0.36\columnwidth, cfbox=red 1pt 0pt]{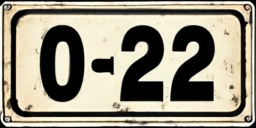}
\includegraphics[width=0.36\columnwidth, cfbox=red 1pt 0pt]{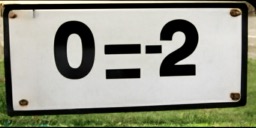}

\vspace{0.5ex}
\textit{"0-6-2"} spelled by ByT5-XL \\[0.3ex]
\includegraphics[width=0.36\columnwidth, cfbox=white 1pt 0pt]{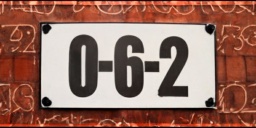}
\includegraphics[width=0.36\columnwidth, cfbox=red 1pt 0pt]{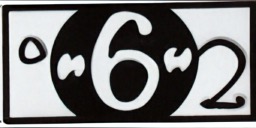}
\includegraphics[width=0.36\columnwidth, cfbox=white 1pt 0pt]{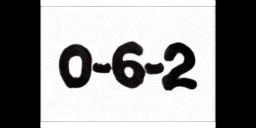}
\includegraphics[width=0.36\columnwidth, cfbox=white 1pt 0pt]{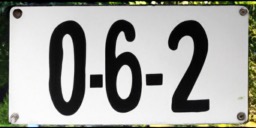}

\vspace{0.5ex}
\textit{"0-6-2"} spelled by ByT5-XXL \\[0.3ex]
\includegraphics[width=0.36\columnwidth, cfbox=red 1pt 0pt]{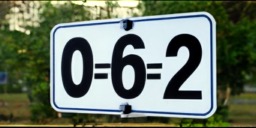}
\includegraphics[width=0.36\columnwidth, cfbox=white 1pt 0pt]{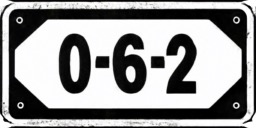}
\includegraphics[width=0.36\columnwidth, cfbox=white 1pt 0pt]{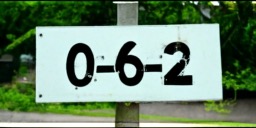}
\includegraphics[width=0.36\columnwidth, cfbox=white 1pt 0pt]{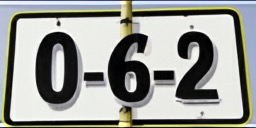}

\vspace{0.5ex}
\textit{"0-6-2"} spelled by Concat \\[0.3ex]
\includegraphics[width=0.36\columnwidth, cfbox=white 1pt 0pt]{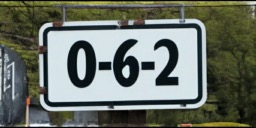}
\includegraphics[width=0.36\columnwidth, cfbox=white 1pt 0pt]{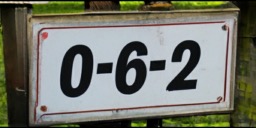}
\includegraphics[width=0.36\columnwidth, cfbox=white 1pt 0pt]{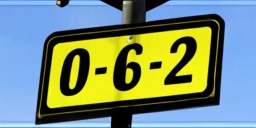}
\includegraphics[width=0.36\columnwidth, cfbox=white 1pt 0pt]{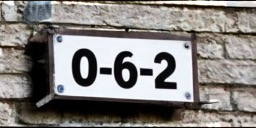}

\caption{Spelling examples from both character-blind (T5) and character-aware (ByT5 and Concat) models. Erroneous spellings are outlined in red. Prompt: \textit{A sign with the word "0-6-2" written on it.}}
\label{fig:samples-062}
\end{figure*}

\begin{figure*}
\centering
\textit{"barratrously"} spelled by T5-XL \\[0.5ex]
\includegraphics[width=0.36\columnwidth, cfbox=red 1pt 0pt]{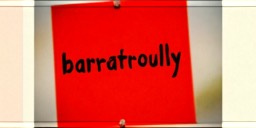}
\includegraphics[width=0.36\columnwidth, cfbox=red 1pt 0pt]{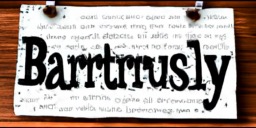}
\includegraphics[width=0.36\columnwidth, cfbox=red 1pt 0pt]{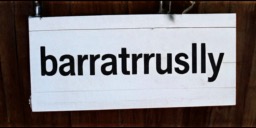}
\includegraphics[width=0.36\columnwidth, cfbox=red 1pt 0pt]{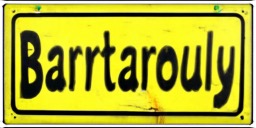}

\vspace{0.5ex}
\textit{"barratrously"} spelled by T5-XXL \\[0.3ex]
\includegraphics[width=0.36\columnwidth, cfbox=red 1pt 0pt]{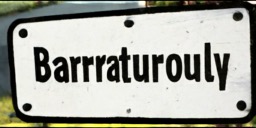}
\includegraphics[width=0.36\columnwidth, cfbox=red 1pt 0pt]{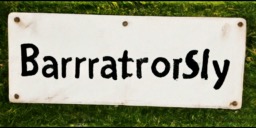}
\includegraphics[width=0.36\columnwidth, cfbox=red 1pt 0pt]{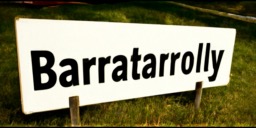}
\includegraphics[width=0.36\columnwidth, cfbox=red 1pt 0pt]{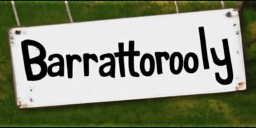}

\vspace{0.5ex}
\textit{"barratrously"} spelled by ByT5-XL \\[0.3ex]
\includegraphics[width=0.36\columnwidth, cfbox=white 1pt 0pt]{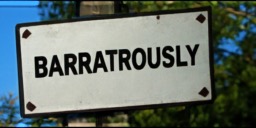}
\includegraphics[width=0.36\columnwidth, cfbox=white 1pt 0pt]{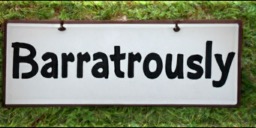}
\includegraphics[width=0.36\columnwidth, cfbox=white 1pt 0pt]{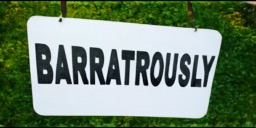}
\includegraphics[width=0.36\columnwidth, cfbox=white 1pt 0pt]{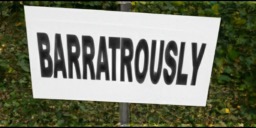}

\vspace{0.5ex}
\textit{"barratrously"} spelled by ByT5-XXL \\[0.3ex]
\includegraphics[width=0.36\columnwidth, cfbox=white 1pt 0pt]{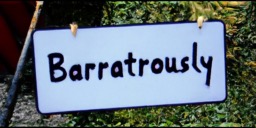}
\includegraphics[width=0.36\columnwidth, cfbox=red 1pt 0pt]{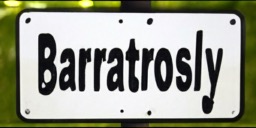}
\includegraphics[width=0.36\columnwidth, cfbox=white 1pt 0pt]{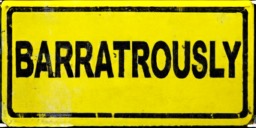}
\includegraphics[width=0.36\columnwidth, cfbox=white 1pt 0pt]{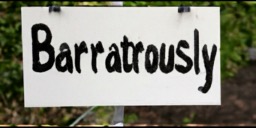}

\vspace{0.5ex}
\textit{"barratrously"} spelled by Concat \\[0.3ex]
\includegraphics[width=0.36\columnwidth, cfbox=white 1pt 0pt]{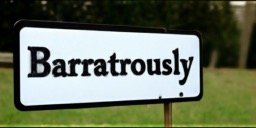}
\includegraphics[width=0.36\columnwidth, cfbox=white 1pt 0pt]{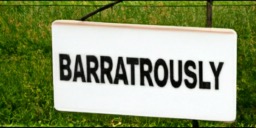}
\includegraphics[width=0.36\columnwidth, cfbox=white 1pt 0pt]{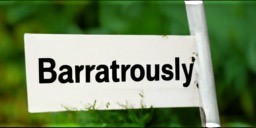}
\includegraphics[width=0.36\columnwidth, cfbox=white 1pt 0pt]{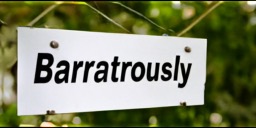}

\caption{Spelling examples from both character-blind (T5) and character-aware (ByT5 and Concat) models. Erroneous spellings are outlined in red. Prompt: \textit{A sign with the word "barratrously" written on it.}}
\label{fig:samples-barratrously}
\end{figure*}

\begin{figure*}
\centering
\textit{"depositories"} spelled by T5-XL \\[0.5ex]
\includegraphics[width=0.36\columnwidth, cfbox=white 1pt 0pt]{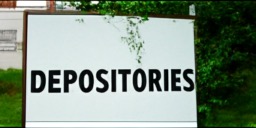}
\includegraphics[width=0.36\columnwidth, cfbox=red 1pt 0pt]{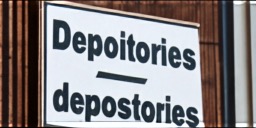}
\includegraphics[width=0.36\columnwidth, cfbox=red 1pt 0pt]{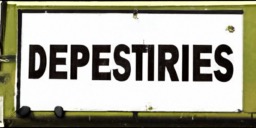}
\includegraphics[width=0.36\columnwidth, cfbox=red 1pt 0pt]{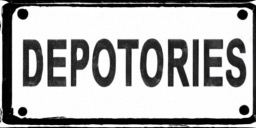}

\vspace{0.5ex}
\textit{"depositories"} spelled by T5-XXL \\[0.3ex]
\includegraphics[width=0.36\columnwidth, cfbox=red 1pt 0pt]{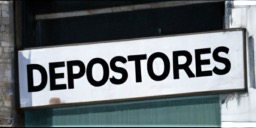}
\includegraphics[width=0.36\columnwidth, cfbox=red 1pt 0pt]{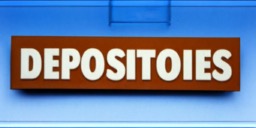}
\includegraphics[width=0.36\columnwidth, cfbox=red 1pt 0pt]{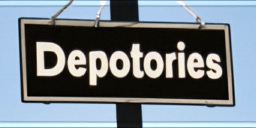}
\includegraphics[width=0.36\columnwidth, cfbox=red 1pt 0pt]{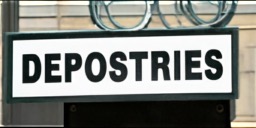}

\vspace{0.5ex}
\textit{"depositories"} spelled by ByT5-XL \\[0.3ex]
\includegraphics[width=0.36\columnwidth, cfbox=white 1pt 0pt]{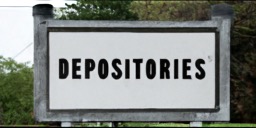}
\includegraphics[width=0.36\columnwidth, cfbox=white 1pt 0pt]{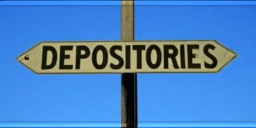}
\includegraphics[width=0.36\columnwidth, cfbox=white 1pt 0pt]{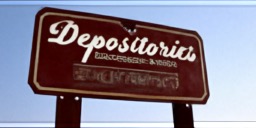}
\includegraphics[width=0.36\columnwidth, cfbox=white 1pt 0pt]{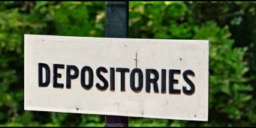}

\vspace{0.5ex}
\textit{"depositories"} spelled by ByT5-XXL \\[0.3ex]
\includegraphics[width=0.36\columnwidth, cfbox=white 1pt 0pt]{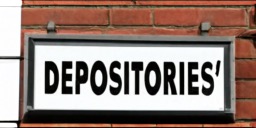}
\includegraphics[width=0.36\columnwidth, cfbox=white 1pt 0pt]{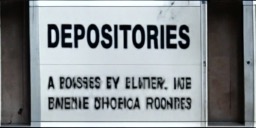}
\includegraphics[width=0.36\columnwidth, cfbox=white 1pt 0pt]{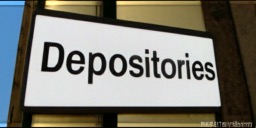}
\includegraphics[width=0.36\columnwidth, cfbox=red 1pt 0pt]{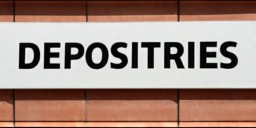}

\vspace{0.5ex}
\textit{"depositories"} spelled by Concat \\[0.3ex]
\includegraphics[width=0.36\columnwidth, cfbox=white 1pt 0pt]{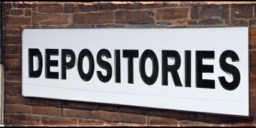}
\includegraphics[width=0.36\columnwidth, cfbox=white 1pt 0pt]{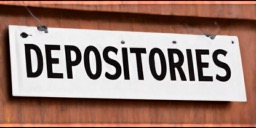}
\includegraphics[width=0.36\columnwidth, cfbox=white 1pt 0pt]{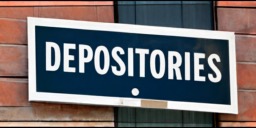}
\includegraphics[width=0.36\columnwidth, cfbox=white 1pt 0pt]{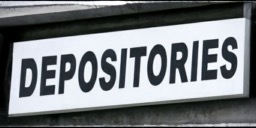}

\caption{Spelling examples from both character-blind (T5) and character-aware (ByT5 and Concat) models. Erroneous spellings are outlined in red. Prompt: \textit{A sign with the word "depositories" written on it.}}
\label{fig:samples-depositories}
\end{figure*}

\begin{figure*}
\centering
\textit{"enceinte"} spelled by T5-XL \\[0.5ex]
\includegraphics[width=0.36\columnwidth, cfbox=red 1pt 0pt]{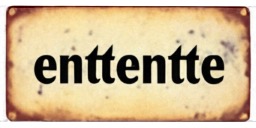}
\includegraphics[width=0.36\columnwidth, cfbox=red 1pt 0pt]{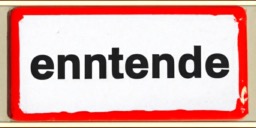}
\includegraphics[width=0.36\columnwidth, cfbox=red 1pt 0pt]{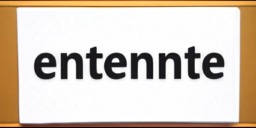}
\includegraphics[width=0.36\columnwidth, cfbox=red 1pt 0pt]{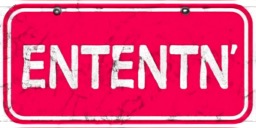}

\vspace{0.5ex}
\textit{"enceinte"} spelled by T5-XXL \\[0.3ex]
\includegraphics[width=0.36\columnwidth, cfbox=red 1pt 0pt]{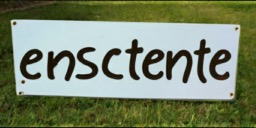}
\includegraphics[width=0.36\columnwidth, cfbox=red 1pt 0pt]{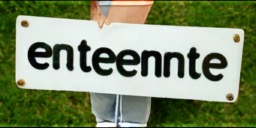}
\includegraphics[width=0.36\columnwidth, cfbox=red 1pt 0pt]{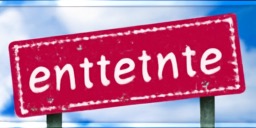}
\includegraphics[width=0.36\columnwidth, cfbox=red 1pt 0pt]{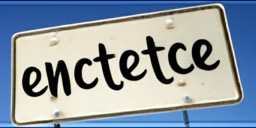}

\vspace{0.5ex}
\textit{"enceinte"} spelled by ByT5-XL \\[0.3ex]
\includegraphics[width=0.36\columnwidth, cfbox=white 1pt 0pt]{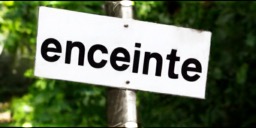}
\includegraphics[width=0.36\columnwidth, cfbox=white 1pt 0pt]{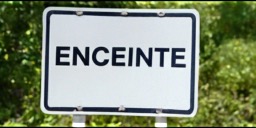}
\includegraphics[width=0.36\columnwidth, cfbox=white 1pt 0pt]{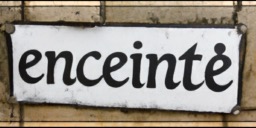}
\includegraphics[width=0.36\columnwidth, cfbox=white 1pt 0pt]{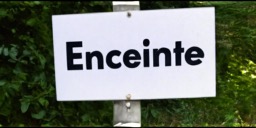}

\vspace{0.5ex}
\textit{"enceinte"} spelled by ByT5-XXL \\[0.3ex]
\includegraphics[width=0.36\columnwidth, cfbox=white 1pt 0pt]{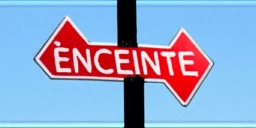}
\includegraphics[width=0.36\columnwidth, cfbox=white 1pt 0pt]{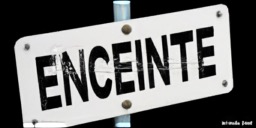}
\includegraphics[width=0.36\columnwidth, cfbox=white 1pt 0pt]{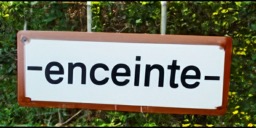}
\includegraphics[width=0.36\columnwidth, cfbox=white 1pt 0pt]{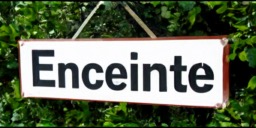}

\vspace{0.5ex}
\textit{"enceinte"} spelled by Concat \\[0.3ex]
\includegraphics[width=0.36\columnwidth, cfbox=white 1pt 0pt]{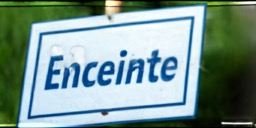}
\includegraphics[width=0.36\columnwidth, cfbox=white 1pt 0pt]{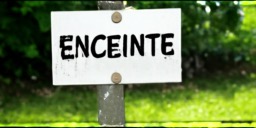}
\includegraphics[width=0.36\columnwidth, cfbox=white 1pt 0pt]{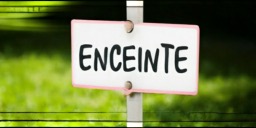}
\includegraphics[width=0.36\columnwidth, cfbox=white 1pt 0pt]{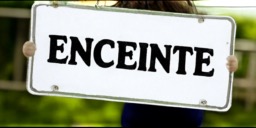}

\caption{Spelling examples from both character-blind (T5) and character-aware (ByT5 and Concat) models. Erroneous spellings are outlined in red. Prompt: \textit{A sign with the word "enceinte" written on it.}}
\label{fig:samples-enceinte}
\end{figure*}

\begin{figure*}
\centering
\textit{"kilopascals"} spelled by T5-XL \\[0.5ex]
\includegraphics[width=0.36\columnwidth, cfbox=red 1pt 0pt]{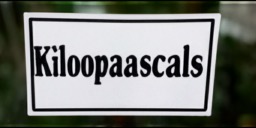}
\includegraphics[width=0.36\columnwidth, cfbox=red 1pt 0pt]{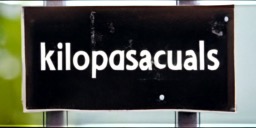}
\includegraphics[width=0.36\columnwidth, cfbox=red 1pt 0pt]{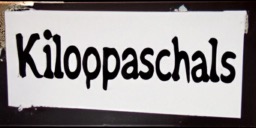}
\includegraphics[width=0.36\columnwidth, cfbox=white 1pt 0pt]{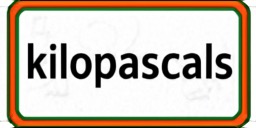}

\vspace{0.5ex}
\textit{"kilopascals"} spelled by T5-XXL \\[0.3ex]
\includegraphics[width=0.36\columnwidth, cfbox=red 1pt 0pt]{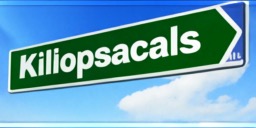}
\includegraphics[width=0.36\columnwidth, cfbox=red 1pt 0pt]{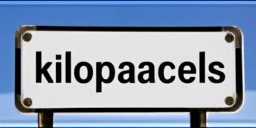}
\includegraphics[width=0.36\columnwidth, cfbox=red 1pt 0pt]{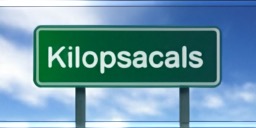}
\includegraphics[width=0.36\columnwidth, cfbox=red 1pt 0pt]{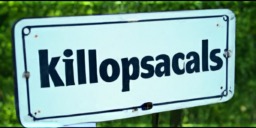}

\vspace{0.5ex}
\textit{"kilopascals"} spelled by ByT5-XL \\[0.3ex]
\includegraphics[width=0.36\columnwidth, cfbox=white 1pt 0pt]{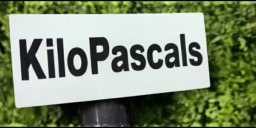}
\includegraphics[width=0.36\columnwidth, cfbox=white 1pt 0pt]{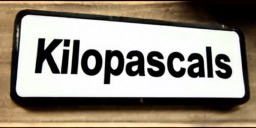}
\includegraphics[width=0.36\columnwidth, cfbox=white 1pt 0pt]{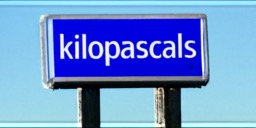}
\includegraphics[width=0.36\columnwidth, cfbox=white 1pt 0pt]{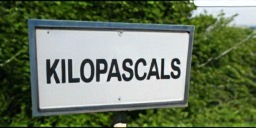}

\vspace{0.5ex}
\textit{"kilopascals"} spelled by ByT5-XXL \\[0.3ex]
\includegraphics[width=0.36\columnwidth, cfbox=white 1pt 0pt]{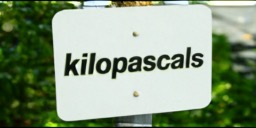}
\includegraphics[width=0.36\columnwidth, cfbox=white 1pt 0pt]{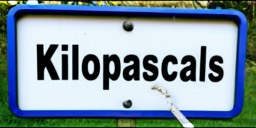}
\includegraphics[width=0.36\columnwidth, cfbox=white 1pt 0pt]{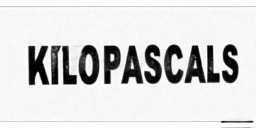}
\includegraphics[width=0.36\columnwidth, cfbox=white 1pt 0pt]{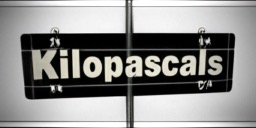}

\vspace{0.5ex}
\textit{"kilopascals"} spelled by Concat \\[0.3ex]
\includegraphics[width=0.36\columnwidth, cfbox=white 1pt 0pt]{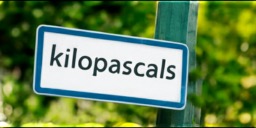}
\includegraphics[width=0.36\columnwidth, cfbox=white 1pt 0pt]{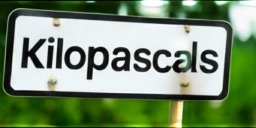}
\includegraphics[width=0.36\columnwidth, cfbox=white 1pt 0pt]{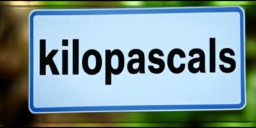}
\includegraphics[width=0.36\columnwidth, cfbox=white 1pt 0pt]{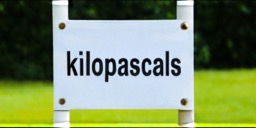}

\caption{Spelling examples from both character-blind (T5) and character-aware (ByT5 and Concat) models. Erroneous spellings are outlined in red. Prompt: \textit{A sign with the word "kilopascals" written on it.}}
\label{fig:samples-kilopascals}
\end{figure*}

\begin{figure*}
\centering
\textit{"rupiahs"} spelled by T5-XL \\[0.5ex]
\includegraphics[width=0.36\columnwidth, cfbox=red 1pt 0pt]{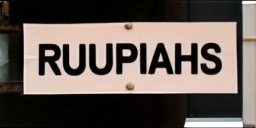}
\includegraphics[width=0.36\columnwidth, cfbox=red 1pt 0pt]{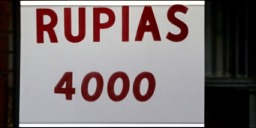}
\includegraphics[width=0.36\columnwidth, cfbox=red 1pt 0pt]{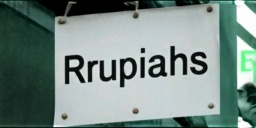}
\includegraphics[width=0.36\columnwidth, cfbox=red 1pt 0pt]{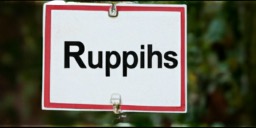}

\vspace{0.5ex}
\textit{"rupiahs"} spelled by T5-XXL \\[0.3ex]
\includegraphics[width=0.36\columnwidth, cfbox=red 1pt 0pt]{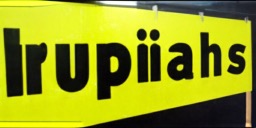}
\includegraphics[width=0.36\columnwidth, cfbox=red 1pt 0pt]{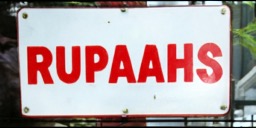}
\includegraphics[width=0.36\columnwidth, cfbox=red 1pt 0pt]{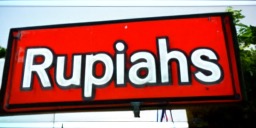}
\includegraphics[width=0.36\columnwidth, cfbox=red 1pt 0pt]{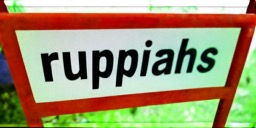}

\vspace{0.5ex}
\textit{"rupiahs"} spelled by ByT5-XL \\[0.3ex]
\includegraphics[width=0.36\columnwidth, cfbox=white 1pt 0pt]{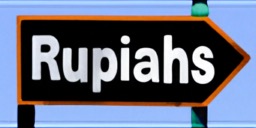}
\includegraphics[width=0.36\columnwidth, cfbox=white 1pt 0pt]{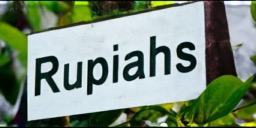}
\includegraphics[width=0.36\columnwidth, cfbox=white 1pt 0pt]{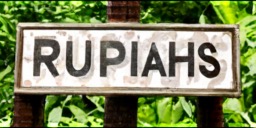}
\includegraphics[width=0.36\columnwidth, cfbox=white 1pt 0pt]{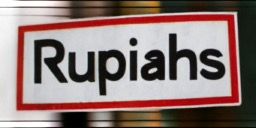}

\vspace{0.5ex}
\textit{"rupiahs"} spelled by ByT5-XXL \\[0.3ex]
\includegraphics[width=0.36\columnwidth, cfbox=white 1pt 0pt]{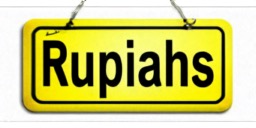}
\includegraphics[width=0.36\columnwidth, cfbox=white 1pt 0pt]{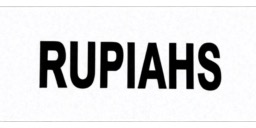}
\includegraphics[width=0.36\columnwidth, cfbox=white 1pt 0pt]{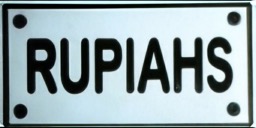}
\includegraphics[width=0.36\columnwidth, cfbox=white 1pt 0pt]{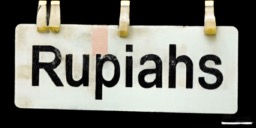}

\vspace{0.5ex}
\textit{"rupiahs"} spelled by Concat \\[0.3ex]
\includegraphics[width=0.36\columnwidth, cfbox=white 1pt 0pt]{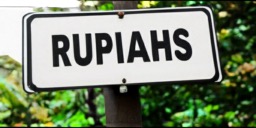}
\includegraphics[width=0.36\columnwidth, cfbox=white 1pt 0pt]{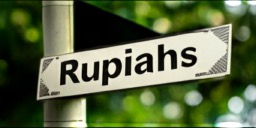}
\includegraphics[width=0.36\columnwidth, cfbox=white 1pt 0pt]{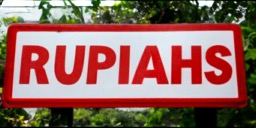}
\includegraphics[width=0.36\columnwidth, cfbox=white 1pt 0pt]{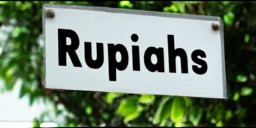}

\caption{Spelling examples from both character-blind (T5) and character-aware (ByT5 and Concat) models. Erroneous spellings are outlined in red. Prompt: \textit{A sign with the word "rupiahs" written on it.}}
\label{fig:samples-rupiahs}
\end{figure*}

\begin{figure*}
\centering
\textit{"Yongchuan"} spelled by T5-XL \\[0.5ex]
\includegraphics[width=0.36\columnwidth, cfbox=white 1pt 0pt]{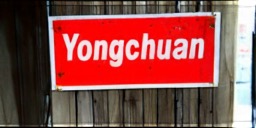}
\includegraphics[width=0.36\columnwidth, cfbox=red 1pt 0pt]{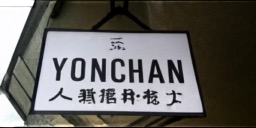}
\includegraphics[width=0.36\columnwidth, cfbox=white 1pt 0pt]{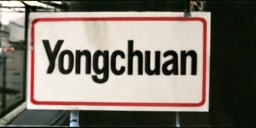}
\includegraphics[width=0.36\columnwidth, cfbox=white 1pt 0pt]{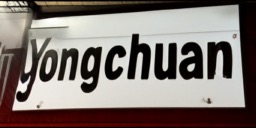}

\vspace{0.5ex}
\textit{"Yongchuan"} spelled by T5-XXL \\[0.3ex]
\includegraphics[width=0.36\columnwidth, cfbox=white 1pt 0pt]{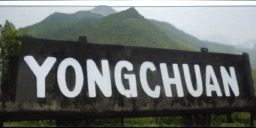}
\includegraphics[width=0.36\columnwidth, cfbox=white 1pt 0pt]{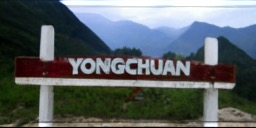}
\includegraphics[width=0.36\columnwidth, cfbox=red 1pt 0pt]{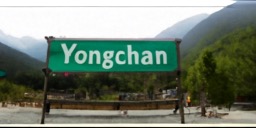}
\includegraphics[width=0.36\columnwidth, cfbox=red 1pt 0pt]{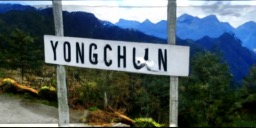}

\vspace{0.5ex}
\textit{"Yongchuan"} spelled by ByT5-XL \\[0.3ex]
\includegraphics[width=0.36\columnwidth, cfbox=white 1pt 0pt]{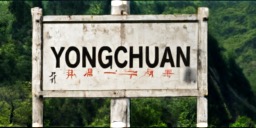}
\includegraphics[width=0.36\columnwidth, cfbox=white 1pt 0pt]{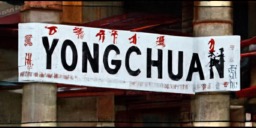}
\includegraphics[width=0.36\columnwidth, cfbox=white 1pt 0pt]{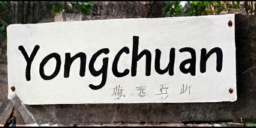}
\includegraphics[width=0.36\columnwidth, cfbox=white 1pt 0pt]{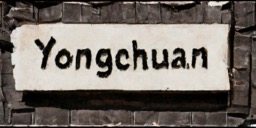}

\vspace{0.5ex}
\textit{"Yongchuan"} spelled by ByT5-XXL \\[0.3ex]
\includegraphics[width=0.36\columnwidth, cfbox=red 1pt 0pt]{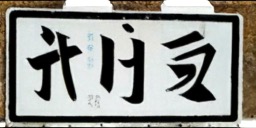}
\includegraphics[width=0.36\columnwidth, cfbox=red 1pt 0pt]{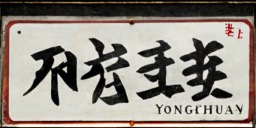}
\includegraphics[width=0.36\columnwidth, cfbox=red 1pt 0pt]{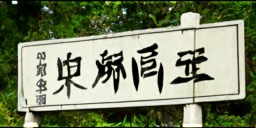}
\includegraphics[width=0.36\columnwidth, cfbox=red 1pt 0pt]{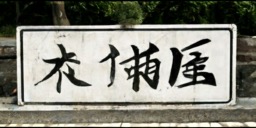}

\vspace{0.5ex}
\textit{"Yongchuan"} spelled by Concat \\[0.3ex]
\includegraphics[width=0.36\columnwidth, cfbox=red 1pt 0pt]{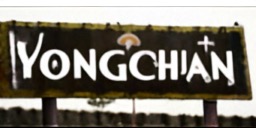}
\includegraphics[width=0.36\columnwidth, cfbox=white 1pt 0pt]{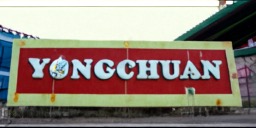}
\includegraphics[width=0.36\columnwidth, cfbox=white 1pt 0pt]{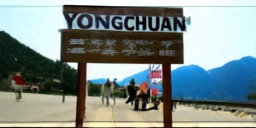}
\includegraphics[width=0.36\columnwidth, cfbox=white 1pt 0pt]{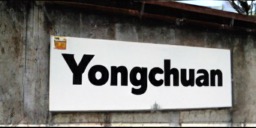}

\caption{Spelling examples from both character-blind (T5) and character-aware (ByT5 and Concat) models. Erroneous spellings are outlined in red. Prompt: \textit{A sign with the word "Yongchuan" written on it.}}
\label{fig:samples-yongchuan}
\end{figure*}

\begin{figure*}
\centering
\textit{"isoaldehyde"} spelled by T5-XL \\[0.5ex]
\includegraphics[width=0.36\columnwidth, cfbox=red 1pt 0pt]{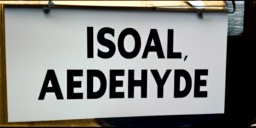}
\includegraphics[width=0.36\columnwidth, cfbox=white 1pt 0pt]{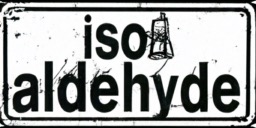}
\includegraphics[width=0.36\columnwidth, cfbox=red 1pt 0pt]{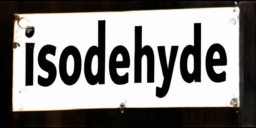}
\includegraphics[width=0.36\columnwidth, cfbox=red 1pt 0pt]{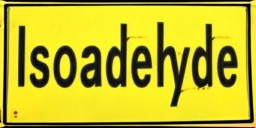}

\vspace{0.5ex}
\textit{"isoaldehyde"} spelled by T5-XXL \\[0.3ex]
\includegraphics[width=0.36\columnwidth, cfbox=red 1pt 0pt]{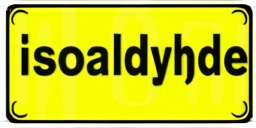}
\includegraphics[width=0.36\columnwidth, cfbox=red 1pt 0pt]{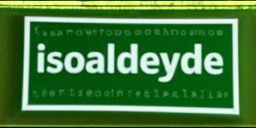}
\includegraphics[width=0.36\columnwidth, cfbox=red 1pt 0pt]{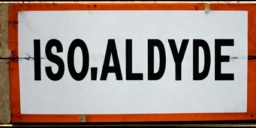}
\includegraphics[width=0.36\columnwidth, cfbox=red 1pt 0pt]{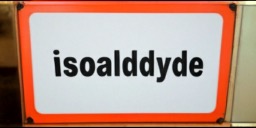}

\vspace{0.5ex}
\textit{"isoaldehyde"} spelled by ByT5-XL \\[0.3ex]
\includegraphics[width=0.36\columnwidth, cfbox=red 1pt 0pt]{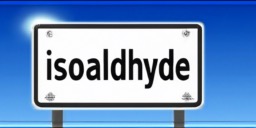}
\includegraphics[width=0.36\columnwidth, cfbox=white 1pt 0pt]{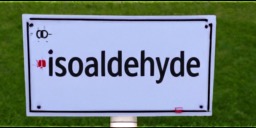}
\includegraphics[width=0.36\columnwidth, cfbox=red 1pt 0pt]{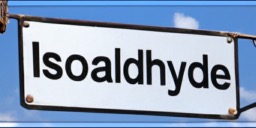}
\includegraphics[width=0.36\columnwidth, cfbox=red 1pt 0pt]{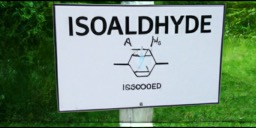}

\vspace{0.5ex}
\textit{"isoaldehyde"} spelled by ByT5-XXL \\[0.3ex]
\includegraphics[width=0.36\columnwidth, cfbox=white 1pt 0pt]{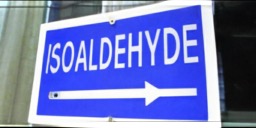}
\includegraphics[width=0.36\columnwidth, cfbox=white 1pt 0pt]{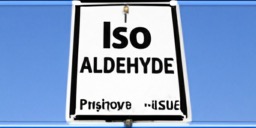}
\includegraphics[width=0.36\columnwidth, cfbox=white 1pt 0pt]{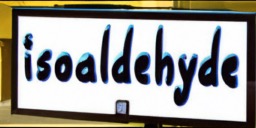}
\includegraphics[width=0.36\columnwidth, cfbox=white 1pt 0pt]{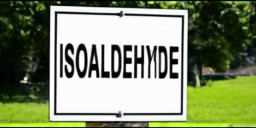}

\vspace{0.5ex}
\textit{"isoaldehyde"} spelled by Concat \\[0.3ex]
\includegraphics[width=0.36\columnwidth, cfbox=white 1pt 0pt]{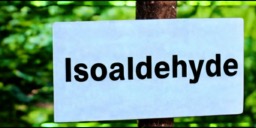}
\includegraphics[width=0.36\columnwidth, cfbox=white 1pt 0pt]{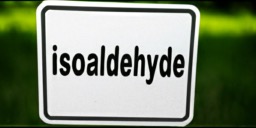}
\includegraphics[width=0.36\columnwidth, cfbox=white 1pt 0pt]{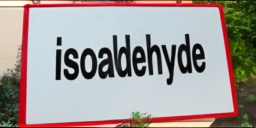}
\includegraphics[width=0.36\columnwidth, cfbox=white 1pt 0pt]{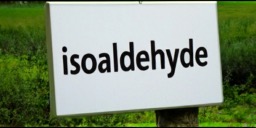}

\caption{Spelling examples from both character-blind (T5) and character-aware (ByT5 and Concat) models. Erroneous spellings are outlined in red. Prompt: \textit{A sign with the word "isoaldehyde" written on it.}}
\label{fig:samples-isoaldehyde}
\end{figure*}

\begin{figure*}
\centering
\textit{"constructivists"} spelled by T5-XL \\[0.5ex]
\includegraphics[width=0.36\columnwidth, cfbox=red 1pt 0pt]{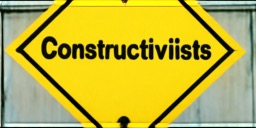}
\includegraphics[width=0.36\columnwidth, cfbox=red 1pt 0pt]{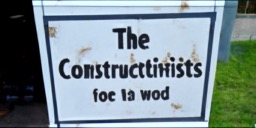}
\includegraphics[width=0.36\columnwidth, cfbox=red 1pt 0pt]{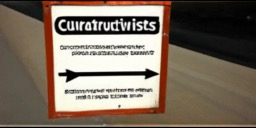}
\includegraphics[width=0.36\columnwidth, cfbox=red 1pt 0pt]{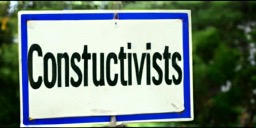}

\vspace{0.5ex}
\textit{"constructivists"} spelled by T5-XXL \\[0.3ex]
\includegraphics[width=0.36\columnwidth, cfbox=red 1pt 0pt]{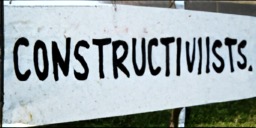}
\includegraphics[width=0.36\columnwidth, cfbox=white 1pt 0pt]{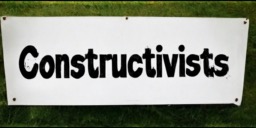}
\includegraphics[width=0.36\columnwidth, cfbox=white 1pt 0pt]{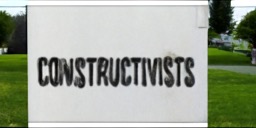}
\includegraphics[width=0.36\columnwidth, cfbox=white 1pt 0pt]{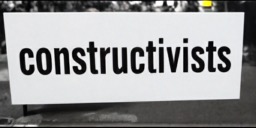}

\vspace{0.5ex}
\textit{"constructivists"} spelled by ByT5-XL \\[0.3ex]
\includegraphics[width=0.36\columnwidth, cfbox=white 1pt 0pt]{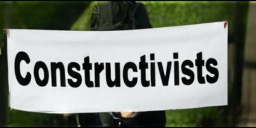}
\includegraphics[width=0.36\columnwidth, cfbox=red 1pt 0pt]{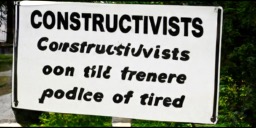}
\includegraphics[width=0.36\columnwidth, cfbox=white 1pt 0pt]{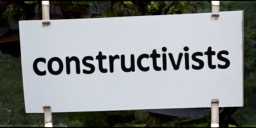}
\includegraphics[width=0.36\columnwidth, cfbox=white 1pt 0pt]{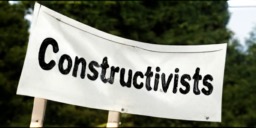}

\vspace{0.5ex}
\textit{"constructivists"} spelled by ByT5-XXL \\[0.3ex]
\includegraphics[width=0.36\columnwidth, cfbox=white 1pt 0pt]{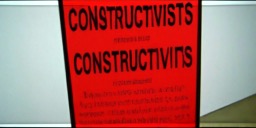}
\includegraphics[width=0.36\columnwidth, cfbox=red 1pt 0pt]{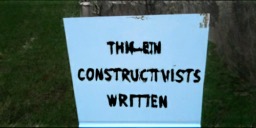}
\includegraphics[width=0.36\columnwidth, cfbox=red 1pt 0pt]{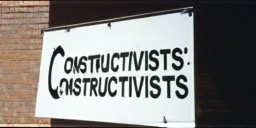}
\includegraphics[width=0.36\columnwidth, cfbox=red 1pt 0pt]{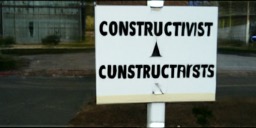}

\vspace{0.5ex}
\textit{"constructivists"} spelled by Concat \\[0.3ex]
\includegraphics[width=0.36\columnwidth, cfbox=white 1pt 0pt]{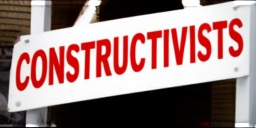}
\includegraphics[width=0.36\columnwidth, cfbox=white 1pt 0pt]{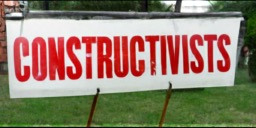}
\includegraphics[width=0.36\columnwidth, cfbox=red 1pt 0pt]{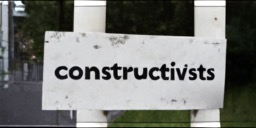}
\includegraphics[width=0.36\columnwidth, cfbox=white 1pt 0pt]{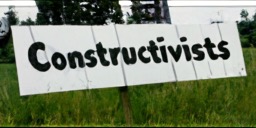}

\caption{Spelling examples from both character-blind (T5) and character-aware (ByT5 and Concat) models. Erroneous spellings are outlined in red. Prompt: \textit{A sign with the word "constructivists" written on it.}}
\label{fig:samples-constructivists}
\end{figure*}

\begin{figure*}
\centering
\textit{"ebike"} spelled by T5-XL \\[0.5ex]
\includegraphics[width=0.36\columnwidth, cfbox=red 1pt 0pt]{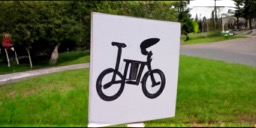}
\includegraphics[width=0.36\columnwidth, cfbox=red 1pt 0pt]{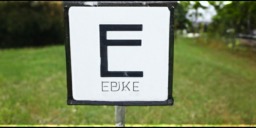}
\includegraphics[width=0.36\columnwidth, cfbox=red 1pt 0pt]{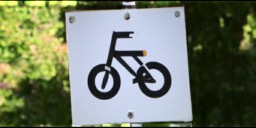}
\includegraphics[width=0.36\columnwidth, cfbox=red 1pt 0pt]{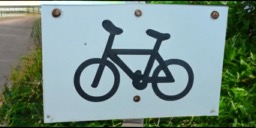}

\vspace{0.5ex}
\textit{"ebike"} spelled by T5-XXL \\[0.3ex]
\includegraphics[width=0.36\columnwidth, cfbox=white 1pt 0pt]{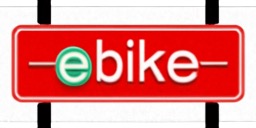}
\includegraphics[width=0.36\columnwidth, cfbox=red 1pt 0pt]{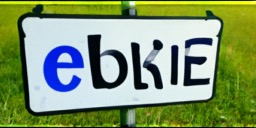}
\includegraphics[width=0.36\columnwidth, cfbox=white 1pt 0pt]{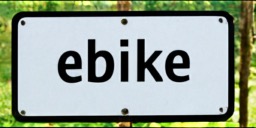}
\includegraphics[width=0.36\columnwidth, cfbox=white 1pt 0pt]{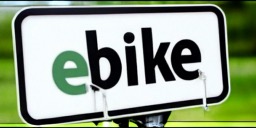}

\vspace{0.5ex}
\textit{"ebike"} spelled by ByT5-XL \\[0.3ex]
\includegraphics[width=0.36\columnwidth, cfbox=red 1pt 0pt]{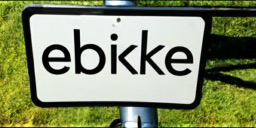}
\includegraphics[width=0.36\columnwidth, cfbox=white 1pt 0pt]{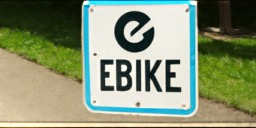}
\includegraphics[width=0.36\columnwidth, cfbox=white 1pt 0pt]{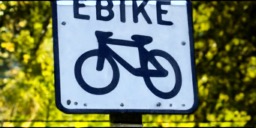}
\includegraphics[width=0.36\columnwidth, cfbox=white 1pt 0pt]{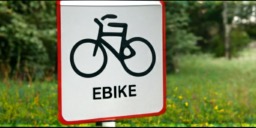}

\vspace{0.5ex}
\textit{"ebike"} spelled by ByT5-XXL \\[0.3ex]
\includegraphics[width=0.36\columnwidth, cfbox=red 1pt 0pt]{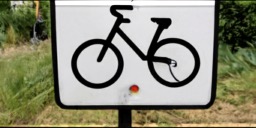}
\includegraphics[width=0.36\columnwidth, cfbox=red 1pt 0pt]{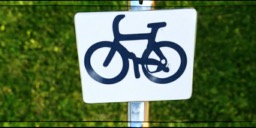}
\includegraphics[width=0.36\columnwidth, cfbox=red 1pt 0pt]{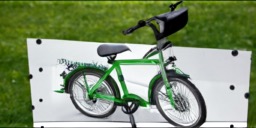}
\includegraphics[width=0.36\columnwidth, cfbox=red 1pt 0pt]{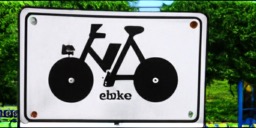}

\vspace{0.5ex}
\textit{"ebike"} spelled by Concat \\[0.3ex]
\includegraphics[width=0.36\columnwidth, cfbox=white 1pt 0pt]{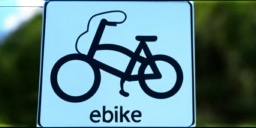}
\includegraphics[width=0.36\columnwidth, cfbox=white 1pt 0pt]{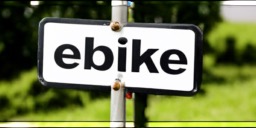}
\includegraphics[width=0.36\columnwidth, cfbox=white 1pt 0pt]{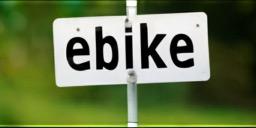}
\includegraphics[width=0.36\columnwidth, cfbox=white 1pt 0pt]{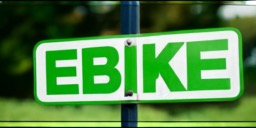}

\caption{Spelling examples from both character-blind (T5) and character-aware (ByT5 and Concat) models. Erroneous spellings are outlined in red. Prompt: \textit{A sign with the word "ebike" written on it.}}
\label{fig:samples-ebike}
\end{figure*}

We show generated image examples from all 5 models (T5-XL, T5-XXL, ByT5-XL, ByT5-XXL, Concat) in Figures  \ref{fig:samples-062} \ref{fig:samples-barratrously}, \ref{fig:samples-depositories}, \ref{fig:samples-enceinte}, \ref{fig:samples-kilopascals}, \ref{fig:samples-rupiahs}, \ref{fig:samples-yongchuan}, \ref{fig:samples-isoaldehyde}, \ref{fig:samples-constructivists}, \ref{fig:samples-ebike}.

Samples are selected based on model's performance. Figures \ref{fig:samples-062} - \ref{fig:samples-rupiahs} are selected words that T5 models tend to get wrong (regardless of ByT5's performance). Figures \ref{fig:samples-yongchuan} - \ref{fig:samples-ebike} are selected words that ByT5 models tend to get wrong (regardless of T5's performance).

\section{DrawText Creative prompts}
\label{sec:appendix_creative_prompts}

We present below $175$ creative prompts targeting rendered text of various lengths: one letter ($10$), one word ($50$), two words ($25$), and three or more words ($90$).

\paragraph{Prompts used in Figure~\ref{fig:creative_prompts}}

\begin{enumerate}
\setlength{\itemsep}{4pt}
\setlength{\parskip}{0pt}
\setlength{\parsep}{0pt}
\begin{footnotesize}

\item	Studio shot of book shelf in the shape of letter G, museum quality, white background.
\item	3-d Letters "DILL" made from dill, studio shot, green background, centered on a page
\item	Word "coffee" made from coffee beans, studio shot.
\item	studio shot multicolored fur in the shape of word "hello", in a furry frame, white background, centered
\item	Photo of a robot lecturer writing the words "Representation Learning" in cursive on a blackboard, with math formulas and diagrams.
\item	studio close-up shot of an antique book with 'knowledge is power' painted in gold on the cover in thick flowing brushed calligraphy
\item	portrait of a parrot is holding a sign with text "no parrots were harmed in the making of this presentation"
\item	A sign that says "Please refrain from arguing with the chimpanzees".

\end{footnotesize}
\end{enumerate}

\paragraph{DrawText Creative prompts: 1 letter}

\begin{enumerate}
\setlength{\itemsep}{4pt}
\setlength{\parskip}{0pt}
\setlength{\parsep}{0pt}
\begin{footnotesize}

\item	Studio shot of book shelf in the shape of letter G, museum quality, white background.
\item	letter "c" made from cactus, high quality photo
\item	Spirograph shape letter M, rainbow lines, white background.
\item	Closeup shot of light magenta, blue and paint brushstrokes of very wide translucent overlapping plastic in the shape of letter F, over white background.
\item	The lowercase letter "b" made out of fire.
\item	Slopy minimal continued line pencil hand drawing of letter Z, white background.
\item	a tower with a huge "w" on the side, from the perspective of a person standing at the base of the tower
\item	3-d letter R made from thin lines connected with dots, white background.
\item	Muted pastel magenta colored paint swirled in white paint in the shape of letter X, globular paint in liquid.
\item	Minimal sculpture of letter W made from light metallic iridescent chrome thin wire, 3-D render, isometric perspective, ultra-detailed, dark background.

\end{footnotesize}
\end{enumerate}

\paragraph{DrawText Creative prompts: 1 word}

\begin{enumerate}
\setlength{\itemsep}{4pt}
\setlength{\parskip}{0pt}
\setlength{\parsep}{0pt}
\begin{footnotesize}

\item	Drops of pastel rainbow colored paint exploding under water in letters "color" shape, pastel rainbow gradient background
\item	3-d Letters "DILL" made from dill, studio shot, green background, centered on a page
\item	Word "coffee" made from coffee beans, studio shot.
\item	studio shot multicolored fur in the shape of word "hello", in a furry frame, white background, centered
\item	Wide lens shot, chunky, organic, colorful, letters "colorful" made from many furry spheres of different sizes, 3-d rendering, centered, studio shot, middle of square canvas
\item	A logo for the company EcoGrow, where the letters look like plants.
\item	a green-colored luxury car with a "green" sticker in the back window
\item	A blackboard with the word "multiplication" written in flowing cursive.
\item	beautiful isometric word "DRAW" entirely made of pencils, soft smooth lighting, pastel colors,  trending on polycount, modular constructivism, blue background, physically based rendering, centered.
\item	transparent water drops exploding under water in the shape of word "water", under water
\item	a drawing of a badger made of mushrooms, with the word "mushroom" written above in glowing letters
\item	a 17th century french baroque painting of a huge female lion, with the word "meow" written in a speech bubble coming from her mouth
\item	a fun and colorful illustration of a waterfall, with the word "waterfall" in the style of a children's book
\item	Letters "VOLUME" fully made from rainbow smoke, black background, centered, sceensaver.
\item	dslr, 3-d word "rainbow" with rainbow fur, white background
\item	a painting of a field of daisies, with the word "danger" written on them in red spray paint
\item	a bottle of hair gel with the label "flawless"
\item	Topographical letters Contour made of a layered paper, muted pastel colors
\item	a logo for the company "brainboost", where the letters look like a brain
\item	a logo for the company "imagine", where the letters look like hands pointing up
\item	A vintage postage stamp showing a painting of the Golden Gate Bridge and the text "California".
\item	a plate of spicy food with the word "spicy" written in flowing cursive
\item	a gold and black logo for the company "moneymoneymoney", which looks like dollar signs
\item	A rendered 3D model of the word "Dependable" made out of granite.
\item	a volcano erupting, with the text "magma" in red
\item	a photo of a prison cell with a window and a view of the ocean, and the word "freedom" painted on the glass
\item	a bowl of alphabet cereal, with the message "smackeroo" written in the bowl with the cereal letters
\item	Studio shot of book shelf in the shape of letters READ, museum quality, white background.
\item	Studio shot of sculpture of text "cheese" made from cheese, with cheese frame.
\item	a landscape of the coyote point national wildlife refuge in arizona, with a coyote sitting on a rock, with the word "coyote" written in sunrise colors
\item	A professional logo for the crypto trading platform "SaltMine".
\item	The word "exquisite" written in modern calligraphy.
\item	A bowl of tomato soup with pasta letters that read "Delicious".
\item	intricate and highly detailed white paper cut out art of a word "SNOW", a storybook illustration, paper cut out, standing in a grotto, made out of white paper, loss of inner self, opening door, hides in the shadows of trees, lithograph, a painting of white silver
\item	3-d letters "dessert" made from desserts, arranged on a plate, studio shot
\item	studio shot of word "BEE" made from bees, white background, in a frame made from bees
\item	The logo for Robotrax, with metallic letters arranged in the shape of a robot.
\item	chunky, organic, colorful, letters "fuzzy" made from many furry spheres of different sizes, 3-d rendering, centered in the frame
\item	photo of a dark cave with the word "crazy" carved into the wall, with a yellow light shining through the cave entrance
\item	a pair of scissors pointing down, and a computer with the word "delete" on the screen
\item	studio shot, word "wow" in script made from rainbow colored fur, in a furry frame, white background, centered
\item	Word "broken" made from broken shattered black glass, centered.
\item	a black and white photo of a saxophone with the word "jazz" written in flowing cursive
\item	Muted pastel multi colored paint swirled in white paint in the shape of letters "swirl", globular paint in liquid
\item	a logo for the company "quantum", where the "q" looks like a lightning bolt
\item	dslr shot of a pair of black and red sneakers with the word "punk" written in white. the background is a dark blue
\item	a logo for the company "diamonds", with a diamond in the shape of a heart
\item	a logo for the company "birthdaypix", where the letters look like birthday candles
\item	a fork with the word "salad" engraved on it in a calligraphic font
\item	3-d word "bricks" with brick texture made from real bricks

\end{footnotesize}
\end{enumerate}

\paragraph{DrawText Creative prompts: 2 words}

\begin{enumerate}
\setlength{\itemsep}{4pt}
\setlength{\parskip}{0pt}
\setlength{\parsep}{0pt}
\begin{footnotesize}

\item	Photo of a robot lecturer writing the words "Representation Learning" in cursive on a blackboard, with math formulas and diagrams.
\item	a sign that reads "no dogs" but with a dog smiling and wagging its tail
\item	a globe with the text "planet earth" in bold letters, with the continents in bright colors
\item	a photo of a sea of roses all around, and a sign in the distance that says "danger: minefield"
\item	giraffe toothbrush made from wood, with the words "giraffe" and "toothbrush" in rainbow color
\item	An airplane flying over a city, with the message "Support Skywriters" written in smoke trails.
\item	A photo of a panda giving a presentation in a large conference room, with text ‘Diffusion Models', in the style of van Gogh
\item	Two llamas dancing the mambo, pointing to a sign that says "Llama Mambo".
\item	A hand painted wooden "Pineapple Club" sign in the shape of a pineapple, hanging outside a bar.
\item	a logo for the company "ethereal media", where the letters look like a painting being created
\item	The cover for the album 'Elusive Interludes' by the band The Melting Snowmen.
\item	A Scrabble board showing the words "optimize" and "pattern".
\item	flowers in a beautiful garden with a text "peace" made by the flowers, with a text "tensions" on the clouds in the sky
\item	a detailed drawing, of words "Vintage lettering", letterism, heavy-gauge filigree, inhabited initials, medium: black pencil, revolver, ecopunk rococo, photo taken of an epic intricate, centered
\item	Bananas arranged on a picnic table to form the message "That's bananas!"
\item	An antique bottle labeled "Energy Tonic".
\item	photo of a helicopter with the text "helicopter tours" on the side landing on a helipad in a valley with a river, trees, and mountains in the background
\item	photo of a sign with "one way"
\item	a sculpture of a brain made from wire and paper, with the words "deep thoughts" written into the material of the brain
\item	a logo for a grocery store chain with the name "grocery land", with the g and the y are made of fruits and vegetables
\item	studio shot of sculpture of text "unlock creativity" made from colorful thin wires
\item	studio shot of a sculpture of a pair of shoes made of colorful wires and the text "unlock creativity"
\item	a vintage image of the las vegas strip with the text "las vegas" in bold block letters
\item	A robot writing "Ethics 101" in chalk on a blackboard.
\item	a yellow saxophone in a rainbow-colored mist with the words "funky mist" that looks like musical clouds of smoke

\end{footnotesize}
\end{enumerate}

\paragraph{DrawText Creative prompts: 3+ words}

\begin{enumerate}
\setlength{\itemsep}{4pt}
\setlength{\parskip}{0pt}
\setlength{\parsep}{0pt}
\begin{footnotesize}

\item	studio close-up shot of an antique book with 'knowledge is power' painted in gold on the cover in thick flowing brushed calligraphy
\item	portrait of a parrot is holding a sign with text "no parrots were harmed in the making of this presentation"
\item	words "Struck by Lightning Twice." made from lightning
\item	a photograph of a field of dandelions with the text "dandelions are the first to go when the lawn is mowed"
\item	a composition of the taj mahal in the center of a gold leaf mandala, with the words "place of honor" centered at the bottom
\item	A poster titled "Quails of North America", showing different kinds of quails.
\item	a cartoon of a cat with a thought bubble saying "this is so weird"
\item	a parrot on a pirate ship, with a parrot wearing a pirate hat, and the caption "i'm the captain now"
\item	Generative art of words "Time is temporary, everything is temporary", viscous smoke made from dots, rivers, graph design, white background.
\item	Studio shot of words "the food is terrible and the portions are too small" made from hotdogs, museum quality, framed photo, white background.
\item	a picture of a powerful-looking vehicle that looks like it was designed to go off-road, with a text saying "i'm a truck, not a car"
\item	a minimalistic version of a forest with a sign saying "help the forest" in the foreground
\item	a map of the world with the text "the world is your oyster" in the middle
\item	cartoon of a dog in a chef's hat, with a thought bubble saying "i can't remember anything!"
\item	A retro coffee ad with the text 'Coffee is what i like'.
\item	different colored shapes on a surface in the shape of words "Life is like a rainbow", an abstract sculpture, polycount, wrinkled, flowing realistic fabric, psytrance, cartography, smooth shading techniques, marble skin, old internet art, camouflage scheme, art », medium poly, smoothened
\item	the view from one end of a bench in a park, looking at the sky, with the text "imagine the outcome" in the sky
\item	a giant shoe, with the caption "shoe for hokey pokey"
\item	A newspaper with the headline "Local pig eats prize pumpkin", and a photo showing the half-eaten pumpkin.
\item	A storefront with "The world's best deli" written on it, centered
\item	Grape vines in the shape of text 'open your mind' sprouting out of a head with flowers and butterflies. DSLR photo.
\item	a plate with a single oyster, with a fork and knife sticking out of the oyster, with a caption that says "oysters for lunch"
\item	dslr portrait of a robot is holding a sign with text "i am not a robot"
\item	Studio shot of words "I like coffee because it gives me the illusion that I might be awake." made from coffee liquid, museum quality, white background.
\item	A hastily handwritten note that says "I'll be back at 4:00" taped to a fridge.
\item	A large recipe book titled "Recipes from Peru".
\item	marquee billboard with "my fear of moving stairs is escalating"
\item	shadow of a stone, taken from the point of view of an ant, with the caption "look at that shadow!"
\item	a pumpkin with a mustache and a monocle and a top hat, with the text "you can get rich too" in a speech bubble
\item	a cartoon of a dog holding a telescope looking at a star with a speech bubble saying "i wonder if there's a dog on that planet"
\item	a blueprint of a house, with a triangle for the roof, a square for the walls, and a rectangle for the floor, and with the message "this house is built on the principles of abstraction"
\item	a sunflower field with a tractor about to run over a sunflower, with the caption "after the sunflowers they will come for you"
\item	text "balloons are flying" made from rainbow balloons, pastel background
\item	the hubble telescope and the milky way, with the text "the universe is a mystery, but we are here to solve it"
\item	a heart with the text "i love you", with the letters "love" made of rainbow colors
\item	studio shot of beautiful textbook with title "how to be a manager of managers", white background
\item	A decorative greeting card that reads "Congratulations on achieving state of the art!"
\item	a painting of a cornfield with the words "feed the nation" in simple letters and colors
\item	A sign that says "Please refrain from arguing with the chimpanzees".
\item	a cartoon of a turtle with a thought bubble over its head with the words "what if there was no such thing as a thought bubble?"
\item	"Fall is here" written in autumn leaves floating on a lake.
\item	a crab sitting on a beach with a surfboard, the sun is a giant orange, and the sky is a rainbow, and the crab is thinking "you are all that matters"
\item	the city of toronto as seen from an airplane, with a giant cn tower in the middle of the frame, with the text "the cn tower" in comic sans
\item	a cartoon of a hippo with a speech bubble saying "i'm a hippo, what do you want?"
\item	a lobster in a suit and tie, holding a microphone, with the caption "lobster says what?"
\item	book with "surgery made easy"
\item	art installation of a chair with the text "i got nothin" carved into the backrest
\item	a painting of a landscape, with a handwritten note that says "this painting was not painted by me"
\item	a picture of a bruised apple with the text "apples are good for you" in a fancy font
\item	A photo of a corgi with a sign that says "I am not a real corgi".
\item	Words "It takes AI and rain to make a rainbow" black background, holography, ((neon colors)), colorful swirly magical ripples, bruh moment, intricate white and gold neon, 3d cg, photorelistic.
\item	a black and white logo on words "Every artist was first an amateur." a white background, a wireframe diagram, generative art, branches growing as hair, tropical reef, trademarks and symbols, in a forest, ios icon, composed of random limbs, stone carving, done in the style of matisse, realms, terminals
\item	picture of two hands, one holding a heart, the other holding a lightning bolt, with the text "love is power"
\item	beautiful photo of the alps, with the caption "the best mountains could do"
\item	a pencil sketch of a tree with the title "nothing to tree here"
\item	a dark forest with a single light in the distance, and the text "i've come to talk with you again"
\item	a circle with the text "infinity makes me happy", in a font that looks like it was written by hand
\item	studio shot of vines in the shape of text 'knowledge is power' sprouting, centered
\item	a photo of a beautiful field of poppies with a sign that says "no photos please"
\item	a grumpy sunflower with a "no solar panels" sign
\item	A meme showing a cat attacking a shoe, with the message "I own your sole".
\item	a test tube with a drop of liquid in it, with the text "we've found water on mars!"
\item	a scene with a city in the background, and a single cloud in the foreground, with the text "contemplate the clouds" in rounded cursive
\item	a picture of a dog and a cat with their heads poking out of a cage with a sign saying "no pets allowed"
\item	a 3d model of a 1980s-style computer with the text "my old habit" on the screen
\item	a mouse with a flashlight saying "i'm afraid of the dark"
\item	A photo of a rabbit sipping coffee and reading a book. The book title "The Adventures of Peter Rabbit" is visible.
\item	clown is holding a paper sign with "Even in hard times there's a possibility to have fun."
\item	newspaper with the headline "aliens found in space" and the text "the truth about everything now challenged"
\item	a dog with a speech bubble with the text "woof woof" and a translation speech bubble with the text "other dogs do vex us"
\item	robot on a butter food processing line, with robot looking dejected, with an overhead red light indicating error, with robot saying "i can't believe it's not butter"
\item	a graffiti art of the text "free the pink" on a wall
\item	a lizard sitting on a baseball field home plate, with the words "made it safe" in a speech bubble
\item	a picture of multiple trees at various stages of development, with the caption "growth is a continuous process"
\item	a purple flower with a crown on its head and a speech bubble that says "i am the purple flower!"
\item	a 1950s-style robot with a giant head and a body shaped like a rocket, with the caption "wow, a real spaceman!"
\item	A professionally designed logo for a bakery called Just What I Kneaded.
\item	Minimal sculpture of word "this is the future" made from light metallic iridescent chrome thin wire, 3-D render, isometric perspective, ultra-detailed, dark background.
\item	pillow in the shape of words "ready for the weekend", letterism, funny jumbled letters, [ closeup ]!!, breads, author unknown, flat art, swedish, diaper-shaped, 2000, white clay, surreal object photography
\item	plant in a fancy pot with a "do not touch" sign on it
\item	a picture of the earth with the words "save the earth" in a circle
\item	scholarly elephant reading a newspaper with the headline "elephants take over the world"
\item	photo of a sign with "having a dog named shark at the beach was a mistake"
\item	photo illustration of the earth being struck by multiple lightning strikes that merge, with the caption "astonishment at the speed of light"
\item	a photo of a fish tank with a fish inside, with the text "tank you for visiting!"
\item	the words "Art is never finished, only continued" in paint splatters on a white background, graffiti art, edge of nothingness love, muddy colors, colored woodcut, beautiful, spectral color
\item	photo of a restaurant "the gas station"
\item	A t-shirt with the message "There is no planet B" written on it.
\item	a close up of a figurine of toothpaste tube, a 3D render, candy pastel, with text "brush your teeth" on the tube
\item	A hand-drawn blueprint for a time machine, with the caption "Time Traveling Device".

\end{footnotesize}
\end{enumerate}

\end{document}